\documentclass[10pt,twocolumn,letterpaper]{article}

\usepackage{wacv}              %

\usepackage[utf8]{inputenc} %
\usepackage[T1]{fontenc}    %
\usepackage{url}            %
\usepackage{booktabs}       %
\usepackage{amsfonts}       %
\usepackage{nicefrac}       %
\usepackage{microtype}      %
\usepackage{xcolor}         %
\usepackage{dblfloatfix}
\usepackage[accsupp]{axessibility}  %

\usepackage{wrapfig}

\usepackage[font=small]{caption}

\usepackage{graphicx}
\usepackage{amsmath}
\usepackage{amssymb}
\usepackage{amsfonts}
\usepackage{algorithm}
\usepackage{algpseudocode}
\usepackage{booktabs}
\usepackage{algorithm}
\usepackage{algpseudocode}
\usepackage{booktabs}
\usepackage[toc,page]{appendix}
\usepackage{subcaption}
\usepackage{wrapfig,lipsum,booktabs}
\usepackage{multirow}

\usepackage{hyperref}
\hypersetup{colorlinks, breaklinks, allcolors=black}
\usepackage[capitalize]{cleveref}
\crefname{section}{Sec.}{Secs.}
\Crefname{section}{Section}{Sections}
\Crefname{table}{Table}{Tables}
\crefname{table}{Tab.}{Tabs.}

\usepackage{xcolor}

\def\wacvPaperID{*****} %
\def\confName{WACV}
\def\confYear{2024}

\begin{document}

\title{Towards More Realistic Membership Inference Attacks on Large Diffusion Models}

\author{\textbf{Jan Dubiński} $^{1,2}$\thanks{Corresponding author: jan.dubinski.dokt@pw.edu.pl} \ \thanks{Equal contribution.}
\quad
\textbf{Antoni Kowalczuk} $^{1\dagger}$
\quad
\textbf{Stanisław Pawlak} $^{1}$
\quad
\textbf{Przemyslaw Rokita} $^{1}$ \\
\quad
\textbf{Tomasz Trzcinski} $^{1,2,3}$
\quad
\textbf{Paweł Morawiecki} $^{4}$ 
\and
$^1$\small Warsaw University of Technology \quad $^2$\small IDEAS NCBR \quad $^3$\small Tooploox \quad $^4$\small Polish Academy of Sciences \quad 
}
\maketitle

\begin{abstract}
     Generative diffusion models, including Stable Diffusion and Midjourney, can generate visually appealing, diverse, and high-resolution images for various applications. These models are trained on billions of internet-sourced images, raising significant concerns about the potential unauthorized use of copyright-protected images. In this paper, we examine whether it is possible to determine if a specific image was used in the training set, a problem known in the cybersecurity community as a membership inference attack. Our focus is on Stable Diffusion, and we address the challenge of designing a fair evaluation framework to answer this membership question. We propose a new dataset to establish a fair evaluation setup and apply it to Stable Diffusion, also applicable to other generative models. With the proposed dataset, we execute membership attacks (both known and newly introduced). Our research reveals that previously proposed evaluation setups do not provide a full understanding of the effectiveness of membership inference attacks. We conclude that the membership inference attack remains a significant challenge for large diffusion models (often deployed as black-box systems), indicating that related privacy and copyright issues will persist in the foreseeable future.
\end{abstract}

\section{Introduction}
\label{sec:intro}

In recent years, there have been rapid advancements in generative modeling techniques within the field of deep learning. Among these, generative diffusion models, particularly those utilising the Stable Diffusion framework, have gained prominence due to their capability to generate high-quality, diverse, and intricate samples. These models hold considerable potential for numerous applications, such as data augmentation, art creation, and design optimization. However, as these models become more widely adopted, addressing the privacy concerns linked to their use is essential.
Recently, Getty Images filed a lawsuit against Stability AI, accusing it of {\it unlawfully copying and processing millions of copyright-protected images}~\cite{gettyimages_lawsuit}. This lawsuit comes on the heels of a separate case Getty lodged against Stability in the United Kingdom, as well as a related class-action lawsuit that California-based artists filed against Stability and other emerging companies in the generative AI sector~\cite{artist_lawsuit}.

\begin{figure}[t!]
  \centering
\includegraphics[width=0.5\textwidth]{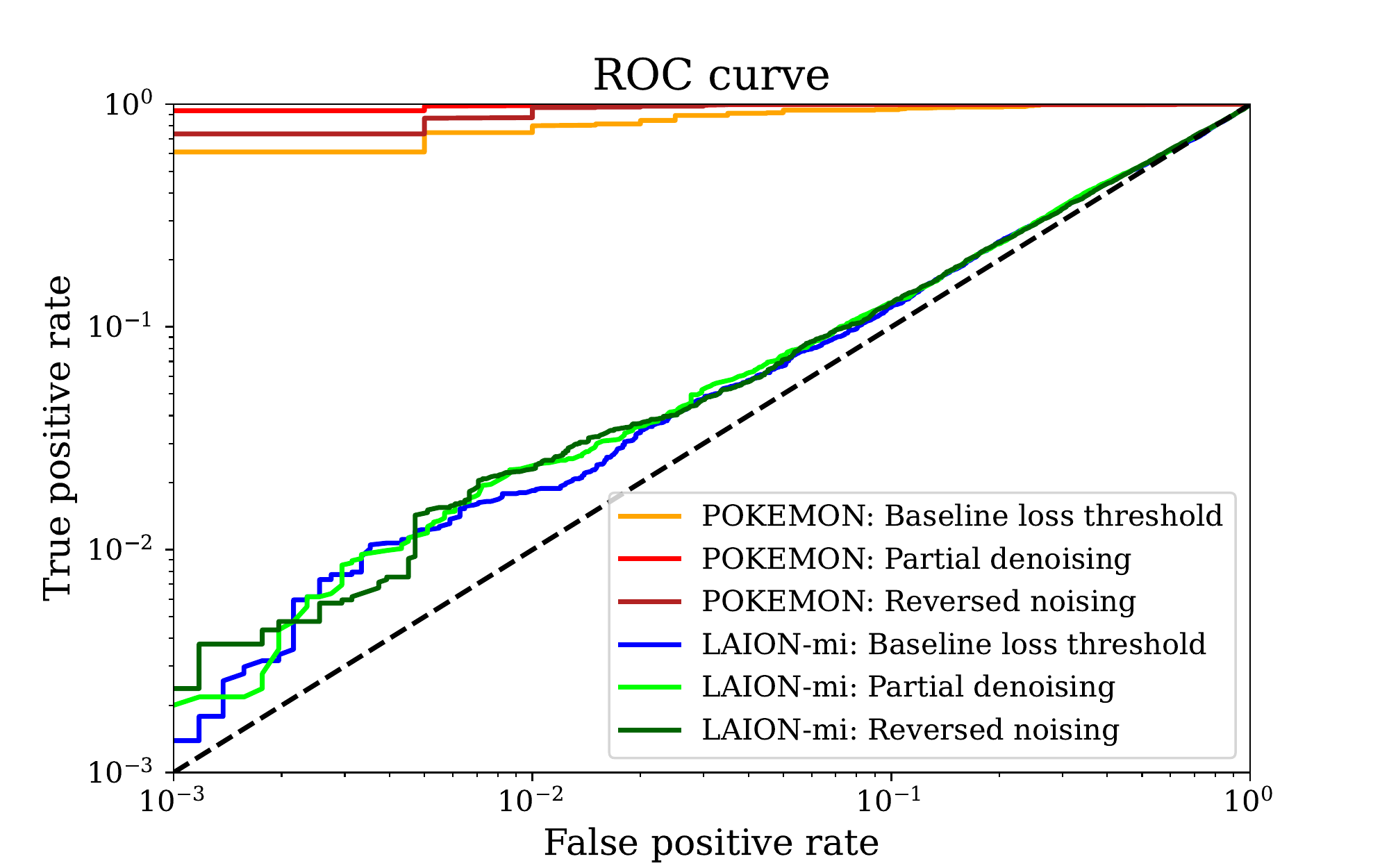}
\caption{ \small \textbf{Pitfalls in the evaluation setting can lead to incorrect conclusions on the effectiveness of membership attacks against large diffusion models such as Stable Diffusion.} An exemplary misleading setup involves finetuning the model on a very small dataset with a low internal variance (such as the POKEMON dataset), which gives a remarkable performance for the selected attacks. However, for the proposed new dataset we observe a drastic performance drop. Our setup does not modify the state-of-the-art Stable Diffusion model but focuses on creating fair membership inference evaluation, possibly close to a real-life usage of the membership attacks.}
    \label{fig:short-a}
\vspace{-5pt}
\end{figure}

One critical issue that arises in this context is determining whether a specific data point was used during the training process of a model. Extracting this information from a model - known as \emph{membership inference attack} - can be crucial in cases where copyrighted or sensitive data are used without permission, leading to potential legal issues.  
Although membership inference attacks have been extensively studied in the context of discriminative models~\cite{carlini2022membership, DBLP:conf/sp/ShokriSSS17, ye2021enhanced,  yeom2018privacy}, the investigation of their effectiveness against generative diffusion models is still in its infancy.

In this paper, we contribute to the understanding of membership inference attacks in large diffusion models, particularly Stable Diffusion. We provide insights into these models, their susceptibility to different membership attacks, and the challenges of their evaluation due to the lack of distinct training and test data. To address these issues, we propose a new dataset for fair and robust evaluation setup. We conduct attacks against Stable Diffusion and assess their effectiveness. Our findings underscore the complexity of data membership inference in large diffusion models. Our main contributions can be summarized as follows:

\begin{itemize}
    
    \item We identify the pitfalls of the existing evaluation of membership inference attacks for large diffusion models. 
    
    \item We provide a new dataset\footnote[1]{\url{https://drive.google.com/drive/folders/17lRvzW4uXDoCf1v_sIiaMnKGIARVunNU}} along with a construction methodology. It allows us to have a more robust evaluation setup for membership inference attacks on the state-of-the-art Stable Diffusion model.
    
    \item With the proposed dataset, we thoroughly evaluate a set of membership inference attacks, which are not prohibitively expensive against Stable Diffusion, including the loss threshold attack and its variants. We also introduce new attacks that focus on modifying the diffusion process to extract more information about membership from the model.
    
\end{itemize}

\section{Background}
\label{sec:background}

\subsection{Diffusion models}

Over the past two years, diffusion models~\cite{sohl2015deep} have emerged as a novel class of generative models, overshadowing Generative Adversarial Networks~\cite{goodfellow2014generative} by achieving state-of-the-art results on numerous benchmarks~\cite{dhariwal2021diffusion} and becoming the core technology behind widely popular image generators such as Stablxe Diffusion~\cite{rombach2022high}, Midjourney~\cite{midjourney}, Runway~\cite{rombach2022high}, Imagen~\cite{saharia2022photorealistic} and DALL-E 2~\cite{ramesh2021zero, ramesh2022hierarchical}.

In essence, \emph{Denoising Diffusion Probabilistic Models}~\cite{ho2020denoising} are probabilistic generative models trained by progressively adding noise to the data and then learning to reverse this process. 

During training, a noised image $x_t \gets \sqrt{a_t} x + \sqrt{1 - a_t} \epsilon$ is produced by adding Gaussian noise $\epsilon \sim \mathcal{N}(0, I)$ to a clean image $x$, with a decaying parameter $a_t \in [0,1]$ such that $a_0 = 1$ and $a_T = 0$. The diffusion model $f_{\theta}$ is trained to remove the noise $\epsilon$ and recover the original image $x$ by predicting the added noise. This is achieved by stochastically minimizing the objective $\frac{1}{N} \sum_i \mathbb{E}_{t, \epsilon} \mathcal{L}(x_i, t, \epsilon; f_{\theta})$, where
\begin{align}
\mathcal{L}(x, t, \epsilon; f_{\theta}) = \lVert \epsilon - f_{\theta}(x_t, t) \rVert_2^2
\label{eq:training_loss}
\end{align}

Despite being trained with a simple denoising objective, diffusion models have shown the ability to generate high-quality images. The process involves sampling a random vector $x_T$ from a normal distribution $\mathcal{N}(0,I)$ and then applying the diffusion model $f_\theta$ to remove the noise from this random image. However, instead of removing all noise at once, the model gradually removes part of the noise iteratively in each generation step.

The final image $x_0$ is obtained from $x_T$ using a noise schedule $\sigma_t$ (dependent on $a_t$), where the model iteratively applies the rule $x_{t-1} = f_{\theta}(x_t, t) + \sigma_t \mathcal{N}(0,I)$ to $\sigma_1=0$. The effectiveness of this process is based on the fact that the diffusion model was trained to denoise images with varying levels of noise.  Applying this iterative generation process with large-scale diffusion models yields results that closely resemble real images.

Certain diffusion models are designed to generate specific types of images by incorporating conditional inputs in addition to the noised image. Class-conditional diffusion models utilise a class label, such as "car" or "plane", to generate a desired image class. Text-conditioned models extend this concept by taking the text embedding of a prompt, such as "a photograph of an astronaut riding a horse in space," which is created by a pretrained language encoder like CLIP~\cite{radford2021learning}.

\subsection{Stable Diffusion}

Stable Diffusion is the largest and most popular open-source diffusion model ~\cite{rombach2022high}. This model is an 890 million parameter text-conditioned diffusion model trained on 2.3 billion images.

Diffusion models can achieve state-of-the-art synthesis results on image data and other applications. However, the optimisation of powerful diffusion models that operate directly in pixel space can consume hundreds of GPU days, and inference can be expensive due to sequential evaluations. To overcome this challenge, the authors of Stable Diffusion~\cite{rombach2022high} propose applying diffusion models in the latent space of powerful pretrained autoencoders. This approach allows for training and inference on limited computational resources while retaining the quality and flexibility of diffusion models.

Formally, given an image $x$, the encoder $E$ encodes $x$ into a latent representation $z = E(x)$, and the decoder $D$ reconstructs the image from the latent, giving $\tilde{x} = D(z) = D(E(x))$. To preprocess conditional information $y$ from various modalities (such as language prompts), the Stable Diffusion framework introduces a domain-specific encoder $\tau_{\theta}$ that projects $y$ to an intermediate representation. Overall,  the Stable Diffusion model is trained by stochastically minimizing the objective $\frac{1}{N} \sum_i \mathbb{E}_{t, \epsilon} \mathcal{L}(z_i, t, \epsilon; f_{\theta})$, where
\begin{align}
\mathcal{L}(z, t, \epsilon; f_{\theta}) = \lVert \epsilon - f_{\theta}(z_t, t,\tau_{\theta}(y) \rVert_2^2
\label{eq:training_loss_sde}
\end{align}

\section{Membership inference attack}

 Membership inference attack \cite{DBLP:conf/sp/ShokriSSS17} answers the question \emph{"was this example in the training set?"}. Currently, two most common approaches are loss based attacks and shadow models.

\subsection{Loss based attacks}
\label{sec:loss_based}
In principle, loss-based membership inference attacks are based on the following simple observation~\cite{yeom2018privacy}. A model training minimises a loss function on a training set, hence we expect the loss to be lower for training samples than for test ones. 
In most cases, such methods treat the attacked model a as white-box, assuming that the attacker has access to the model, its source code and trained weights. This assumption is often not met in practice, as API-based generative machine learning services such as Midjourney~\cite{midjourney} increase in popularity.
In general, methods based solely on the analysis of the loss of the model are less effective than methods utilising shadow models~\cite{carlini2023extracting_diffusion, carlini2022membership}.

\subsection{Shadow models}
\label{sec:shadow_models}
Membership inference attacks based on \emph{shadow models} involve creating multiple models that imitate the behaviour of the target model, but whose training datasets are known to the attacker. By studying the labelled inputs and outputs of these shadow models, researchers can gain insight into the target model's behaviour and develop attacks that can exploit its vulnerabilities. For diffusion models~\cite{carlini2023extracting_diffusion} introduced membership inference attacks called LiRA. This approach involves training a collection of shadow models on random subsets of the training dataset. Once the shadow models have been trained, LiRA computes the loss for each example under each shadow model. By analysing the distribution of losses, LiRA can then determine whether a given example belongs to the training dataset or not. 
Although shadow models have proven to be a powerful tool in the development of membership inference attacks, this approach has its own disadvantages. Such methods are computationally very expensive, as they require the training of multiple copies of the target model. In particular, for large diffusion models such as Stable Diffusion, the cost of developing multiple shadow models is in practice too high.

\section{Attack Challenges and Pitfalls for Large Diffusion Models}
\label{sec:attack_challenges}
\textbf{Lack of nonmembers} To perform and evaluate a \textit{membership inference attack} we need two sets: members and nonmembers. Typically, member data samples are drawn from the training set, whereas nonmembers from the test set. Unfortunately, for the Stable Diffusion model, we cannot follow this approach. The original Stable Diffusion model was trained on the data from the LAION-2B EN dataset, a subset of the LAION-5B dataset~\cite{schuhmann2022laion5b}. Since the dataset is huge (more than 2 billion images) and was not divided into a test and training set, nonmember samples are not easily available. 

\textbf{Shadow models cost} As stated in Section \ref{sec:shadow_models} the shadow models are computationally expensive. The method requires training several dozens of models from scratch. For huge models, such as Stable Diffusion, this approach is practically infeasible. In particular, the cost of training a single Stable Diffusion model is estimated at 600,000\$. Moreover, it would take 80.000 A100 GPU-hours to complete the training.

\textbf{Pitfall 1: Evaluation based on fine-tuning} In \cite{duan2023diffusion} authors propose to tackle the lack of nonmembers by fine-tuning the Stable Diffusion on a dataset that was not used for training the original model. However, it has been demonstrated that fine-tuning the model can easily lead to overfitting~\cite{huggingface_tex_to_img_ft}. As shown in \cite{carlini2023extracting_diffusion}, better diffusion models are more vulnerable to membership inference attacks: the quality of the generated samples is proportional to the success rate of the attack. This is especially the case for models that overfit to the limited training data during fine-tuning~\cite{yeom2018privacy}, leading to inflated performance and misleading conclusions.

\textbf{Pitfall 2: Distribution mismatch between members and nonmembers} Another approach to dealing with the absence of a natural nonmember dataset is to draw samples from a dataset similar to the training data that was not actually used in the training. However, for a fair evaluation of membership inference attacks, it is important that the members and nonmembers share the same feature distribution. If these two groups can be easily distinguished based on feature distribution mismatch, then there is a high risk that the attack method will learn to distinguish between the features of the two data groups~\cite{rejectedICLR2023} instead of the behaviour of the model on these two sets.

\section{The LAION-mi dataset}
\label{sec:laion_mi}
\begin{figure}[t!]
  \centering
\includegraphics[trim={0cm 0cm 0cm 0cm},clip, width=0.4\textwidth]{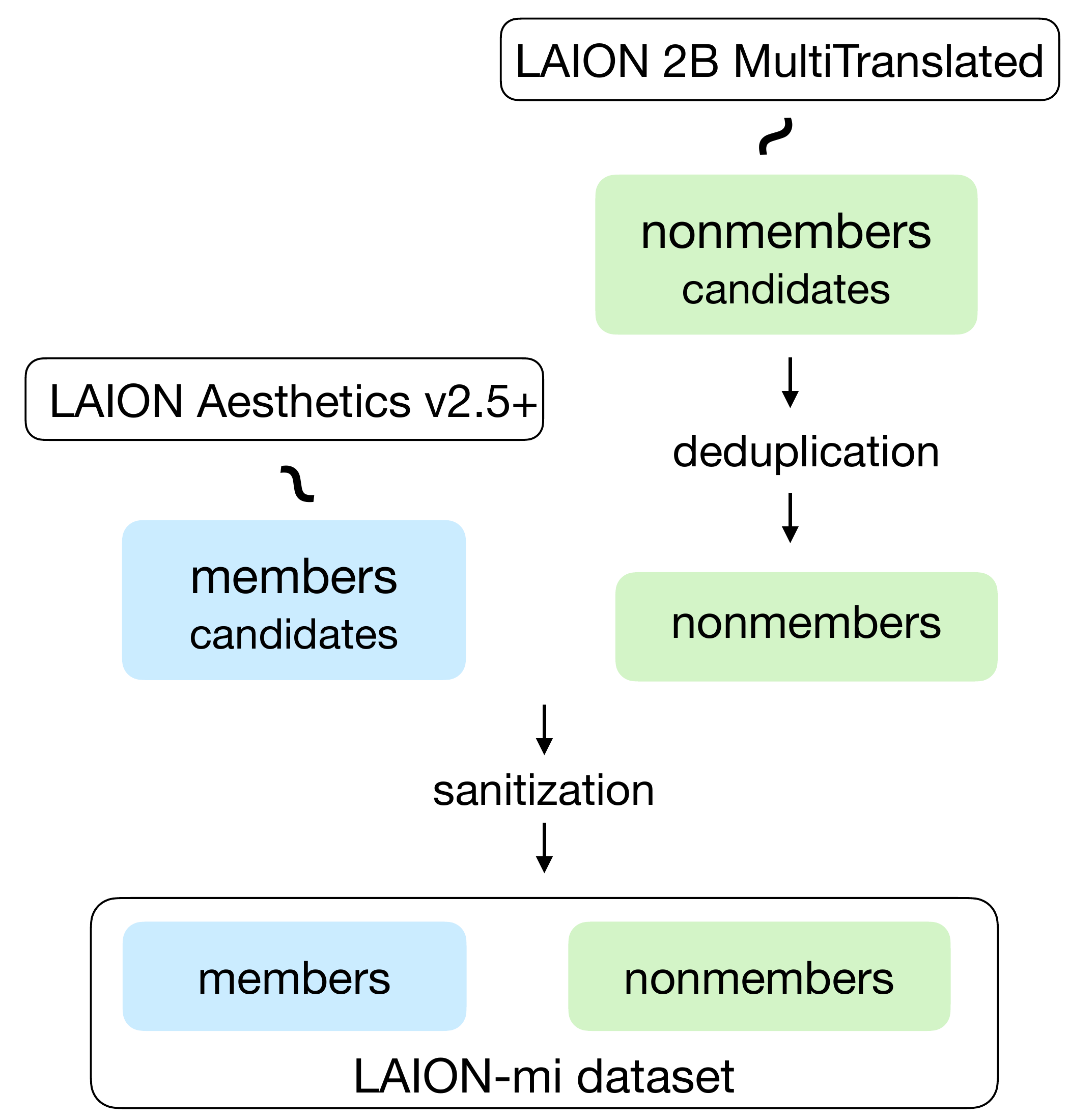}
\caption{ \small \textbf{A general scheme of constructing LAION-mi dataset.} First, members and nonmembers are sampled from LAION Aesthetics v2 5+ and LAION-2B Multi Translated datasets, respectively. Then, we remove nonmember that are duplicates of samples from the member set. Finally, to ensure that the distribution of member samples is indistinguishable from nonmembers distribution, we execute an extensive sanitization algorithm on the member set.}
\label{fig:general_scheme}
\end{figure}

To address the absence of nonmembers samples for the Stable Diffusion model we propose a new dataset consisting of members and nonmembers called LAION-mi. This new dataset aims to facilitate a realistic evaluation of membership inference attacks against large diffusion models. We do not fine-tune nor modify the Stable Diffusion model in any other way, in order to avoid the first pitfall from Sec. \ref{sec:attack_challenges}. To mitigate the second pitfall, we apply a sanitization process on the member set. Figure \ref{fig:general_scheme} shows a general scheme of how the LAION-mi dataset is constructed.

\subsection{Sources of members and nonmembers sets in LAION-mi}
\textbf{Members} Stable Diffusion-v1.4 was trained on all data points from the LAION Aesthetics v2 5+ dataset (see Appendix \ref{sec:appendix_sd14} for details), so all samples from this dataset can serve as member candidates for our new dataset.

\textbf{Nonmembers} To obtain the nonmembers set, we use LAION-2B Multi Translated dataset~\cite{laion_2b_multi_translated}. This dataset is created by LAION-5B authors from LAION-2B Multi, a subset of LAION-5B, but unlike LAION-2B EN, LAION-2B Multi consists of samples which descriptions are in other languages (not English). LAION-2B Multi Translated is obtained by translating these descriptions to English. Since SD-v1.4 was fine-tuned using samples with an aesthetic score above 5 (obtained by LAION-Aesthetics\_Predictor V2~\cite{aesthetic_predictor}), to build our nonmembers set we also use samples with aesthetic score above 5. The score is precomputed by the LAION-2B Multi Translated dataset authors.

\subsection{Adapting members and nonmembers sets to ensure the validity of evaluation setting.}
 As mentioned before, for a fair evaluation of the membership inference attack we should ensure that the underlying data distribution is the same for the member and nonmember samples. We solve it by introducing the adaptation step for members and nonmembers subsets constituting LAION-mi dataset. During the adaptation, we first deduplicate the nonmembers set (Sec. \ref{sec:deduplication}) and then filter the member set using sanitization process (Sec. \ref{sec:sanitization}). Finally, we obtain members and nonmembers sets which have the same, indistinguishable underlying distribution.

\subsection{Deduplication}
\label{sec:deduplication}

\textbf{Member samples in the nonmembers source dataset}
Duplicate samples are images present in a dataset more than once. It has been shown \cite{webster2023deduplication} that LAION-2B EN contains approximately 30\% duplicates when it comes to the image data. We can expect that the whole LAION-5B contains samples, which are present both in LAION-2B EN and LAION-2B Multi Translated. These samples can potentially contaminate LAION-mi nonmembers subset with members samples and therefore compromise the fairness and correctness of our solution. The following procedure aims to address this issue. 

\textbf{Solution}
In order to obtain the nonmembers set free of contamination by members samples we perform two-step deduplication. The first step aims to propose a set of duplicate candidates for each nonmember sample. The second step filters out nonmembers samples, which we suspect have a duplicate in the members source dataset. We end up with the clean nonmembers set.

\textbf{Duplicate candidates} We need to somehow obtain the samples from the LAION-2B EN dataset which are the most similar to the given sample from the nonmember source. We achieve this by querying the Clip Retrieval Client \cite{beaumont-2022-clip-retrieval}. This service searches through the LAION-5B dataset and returns the requested amount of samples that are the most similar to the input image. In our approach, we obtain up to 40 duplicate candidates per nonmember sample. One important limitation of this service is that it doesn't distinguish between subsets of LAION-5B. LAION-5B is split into LAION-2B EN and LAION-2B Multi using the CLD3 \cite{vo2019language} language identifier on the captions of images. Because we care only about the duplicates from the LAION-2B EN, we need to check if the returned candidate is from this dataset. We use CLD3 to check if the given candidate is from the LAION-2B EN dataset. Only samples from the LAION-2B EN dataset are considered duplicate candidates.

\textbf{Duplicates detection and filtering}
When it comes to filtering out the duplicates we propose the following approach: we define the distance metric between the nonmember and its duplicate candidate, and then if the distance is below some threshold we decide that this sample has a duplicate in the members set, effectively discarding it from the final nonmember set.

\begin{figure}[h!]
    \begin{subfigure}{0.23\textwidth}
        \centering
        \includegraphics[width=\textwidth]{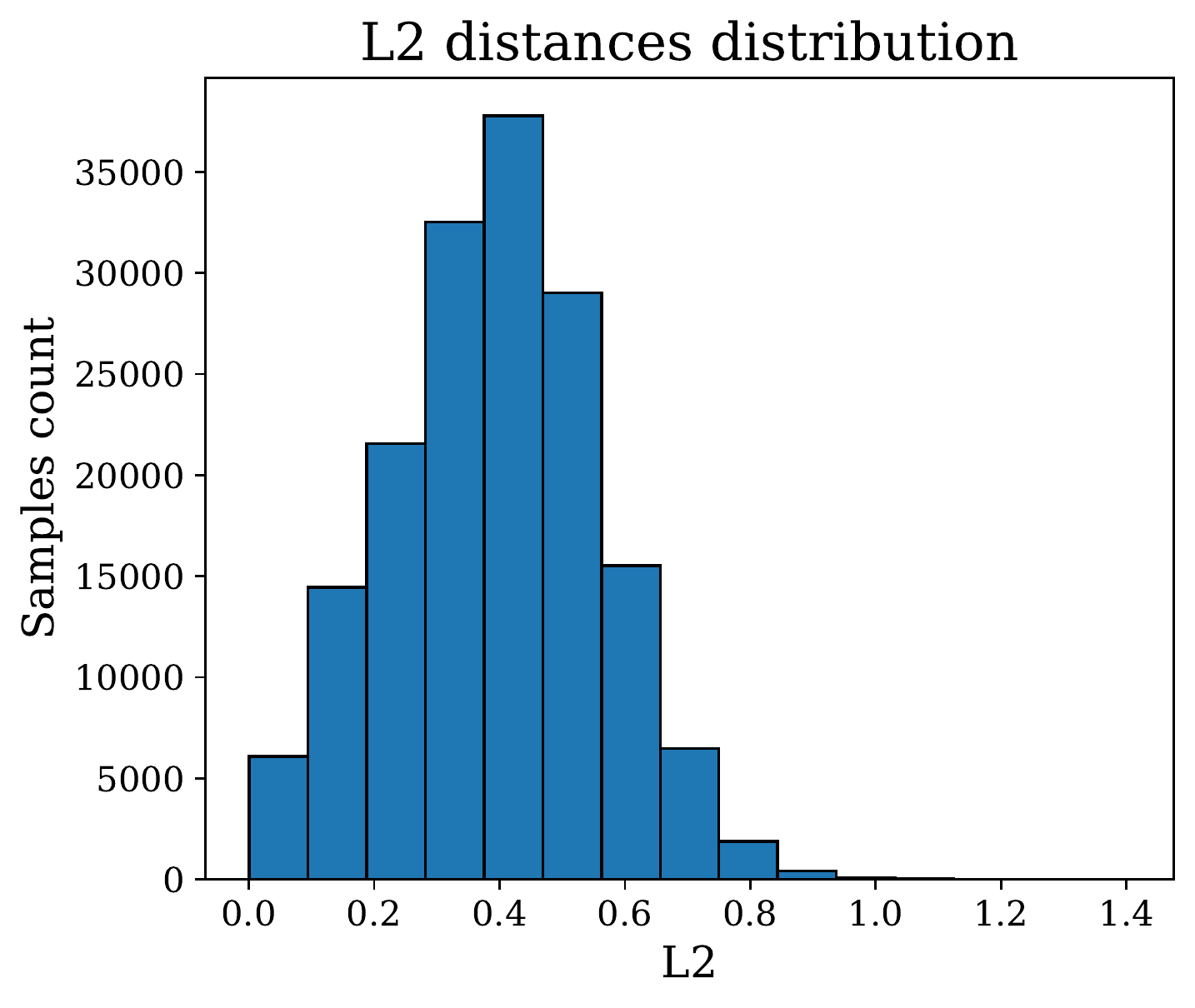}
        \caption{Distribution of L2 distances.\newline}
        \label{fig:l2-dist}
    \end{subfigure}
    \begin{subfigure}{0.23\textwidth}
        \centering
        \includegraphics[width=\textwidth]{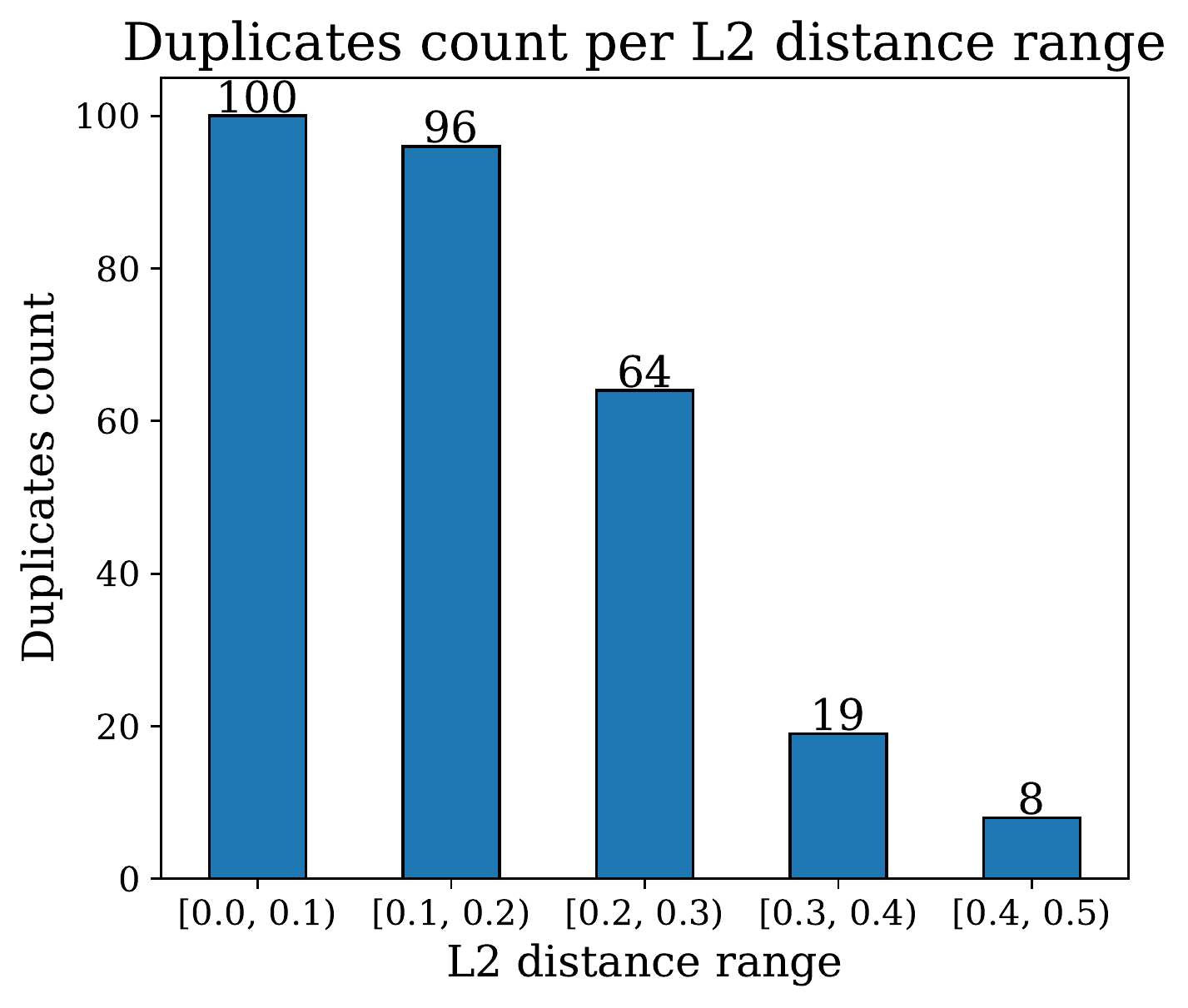}
        \caption{Duplicates count per 100 samples for L2 distances buckets.}
        \label{fig:l2-dist-vs-duplicates}
    \end{subfigure}
    \caption{ \small The distribution of the L2 distances between duplicate candidates and the original images approximately follow a normal distribution, with a mean around $0.4$ (left). For increasing L2 threshold value the duplicates count decreases sharply, with the interval $[0.4, 0.5)$ containing approximately $8\%$ duplicates and $92\%$ non-duplicates.}
    \label{fig:l2-dist-full-plot}
\end{figure}

Firstly, for each sample, we calculate the L2 distances of CLIP image embeddings between the sample and all of its duplicate candidates. Then the final duplicate candidate for each sample is the one with the lowest L2 distance score. We then end up with the approximately normal distribution of L2 distances (see Fig. \ref{fig:l2-dist}).

We then pick the threshold below which we reject samples and mark them as duplicates. Our goal here is to filter out as many duplicates as possible, to avoid contamination of the nonmembers set with members samples. At the end of this process, we have less than $1\%$ of the duplicates by setting this threshold at $0.5$, using \textit{the rule of three} \cite{rule_of_three}. 
We manually confirm it by sampling random $300$ samples and checking for the duplicates, without finding any. For the threshold of $0.5$ we reject approximately $75\%$ of nonmembers as having a duplicate in the members source dataset. It is a really conservative approach, but as we show next, it is necessary in order to achieve the cleanest nonmembers set possible. 

To further confirm that we pick the correct threshold we perform a manual analysis of the duplicates ratio in different L2 score intervals and show the results in Figure \ref{fig:l2-dist-vs-duplicates}. Since our goal is to make the cleanest nonmembers dataset possible, we pick a threshold of $0.5$ to avoid duplicates.

\subsection{Sanitization}
\label{sec:sanitization}

\textbf{Differences between sets}
As we have stated before, one of the most important challenges of membership attack evaluation is ensuring that the member and nonmember samples come from the most similar distribution possible, in our case both images and their descriptions coming from these subsets should be indistinguishable from each other. However, our source of nonmembers is obtained by translating captions from LAION-2B Multi to English using machine translation\footnote[2]{Facebook's M2M100 1.2B model~\cite{fan2020englishcentric}}. Therefore, we expect the distribution of captions' CLIP~\cite{radford2021learning} embeddings to be different for members and nonmembers sets. We need to address this issue to not fall into the second pitfall, which we line up in the Sec. \ref{sec:attack_challenges}.

\begin{figure*}[!ht]
\centering
    \begin{subfigure}{0.32\textwidth}
        \centering
        \includegraphics[trim={0cm 0cm 0cm 1cm},clip ,width=\textwidth]{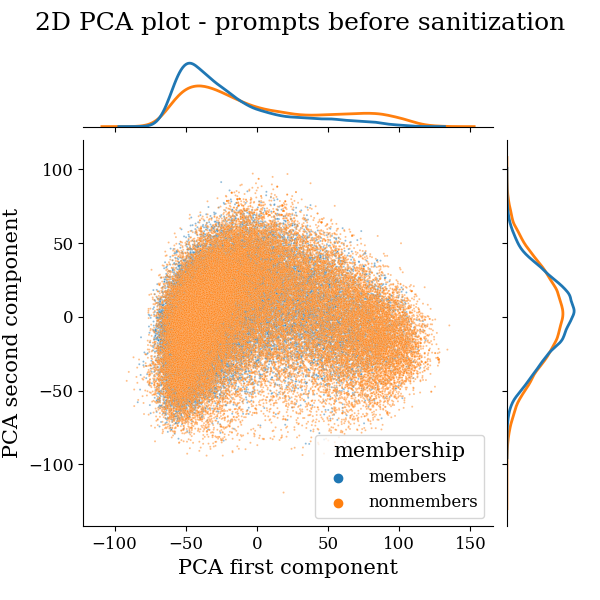}
        \caption{ Prompts embeddings \\ before sanitization}
        \label{fig:prompts_before}
    \end{subfigure}
    \begin{subfigure}{0.32\textwidth}
        \centering
        \includegraphics[trim={0cm 0cm 0cm 1cm},clip, width=\textwidth]{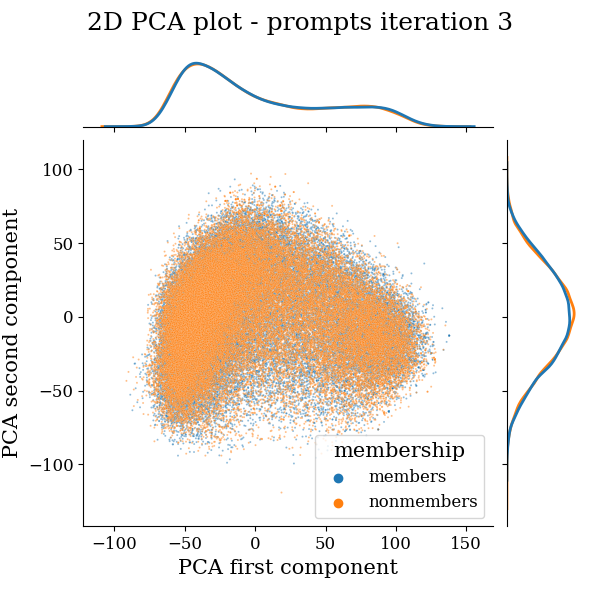}
        \caption{  Prompts embeddings after \\ three iterations}
        \label{fig:prompts_it_3}
    \end{subfigure}
    \begin{subfigure}{0.32\textwidth}
        \centering
        \includegraphics[trim={0cm 0cm 0cm 1cm},clip, width=\textwidth]{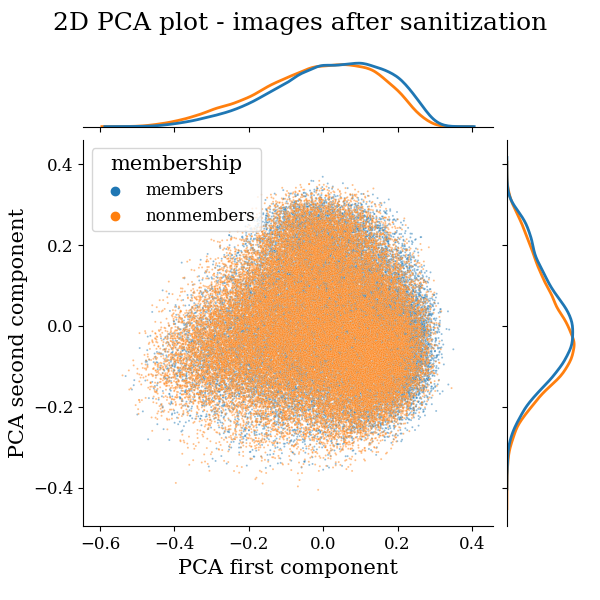}
        \caption{ Image embeddings after\\ three iterations}
        \label{fig:images_it_3}
    \end{subfigure}
    \caption{ \small \textbf{Sanitization effect on prompts and image embeddings distribution of the members and nonmembers sets.} Figure \ref{fig:prompts_before} shows that there is a significant difference between prompts embeddings of nonmembers and members samples before the sanitization process. After three iterations of our sanitization Algorithm \ref{alg:sanitization} these distributions match closely -- Figure \ref{fig:prompts_it_3}. Despite the fact that the algorithm uses only prompts embeddings, we observe in Figure \ref{fig:images_it_3} that image embeddings distributions are also well aligned after the third iteration of our algorithm. We suspect that it is due to the close match between images and their descriptions so that aligning text distributions leads to aligning image distributions as well.}
    \label{fig:pca_sanitization}
\end{figure*}

\textbf{Assessment approach} In order to assess the magnitude of this problem and the efficacy of our sanitization approach we use three metrics. All are based on the CLIP embeddings of the descriptions and images. The metrics are as follows:
\begin{itemize}
    \item Fréchet Inception Distance (FID)~\cite{heusel2018gans}: in order to compare the resulting metric we compute it for two cases: internal (between two random samples of 10k examples from the same set) and comparative (between 10k random members and 10k random nonmembers). If the difference between these two is significant, we assume that there is a mismatch between distributions.
    \item Visual analysis of PCA 2D projection: we use PCA decomposition on the embeddings to project them into a 2D space using scikit-learn \cite{sklearn_api} implementation. The mismatch in the underlying distributions should be indicated by a mismatch of the distributions of the PCA components of the projected embeddings.
    \item Classification: we train a binary classifier in order to distinguish between embeddings of the descriptions. High accuracy indicates a significant difference between the two sets.
\end{itemize}
\textbf{Scale of the problem} Our experiments confirm the seriousness of the issue. Firstly, we observe that FID for the comparative case is way greater than for any of the internal cases, see Tab. \ref{tab:m_vs_nm_fid}. Secondly, visual analysis of embeddings projected to 2D space using PCA, Fig. \ref{fig:prompts_before}) confirms our concerns that these text embeddings are in fact different. Finally, a simple logistic regression model separates prompt embeddings with a $90\%$ accuracy.

\begin{table}[b!]

\begin{sc}
  \begin{center}
  \caption{\small \textbf{FID comparison for 10k samples.} For text data we calculate FID for CLIP embeddings. To calculate FID for image data we first resize images to 512x512 and then use Pytorch FID implementation~\cite{Seitzer2020FID}. To calculate internal FID we divide the dataset into 2 equal random subsets, each of 10k samples.}
  \scalebox{1}{
    \small
  \begin{tabular}{@{}lcc@{}}
    \toprule
& \multicolumn{2}{c}{FID} \\
Data subset & text & images  \\
    \midrule
    Members internal - random & 9.84 & 7.00 \\
    Members internal - sanitized & 9.77 & 7.06 \\
    Nonmembers internal & 9.73 & 7.01 \\
    \midrule
    Comparative - random & 66.43 & 13.90 \\
    Comparative - sanitized & \textbf{13.54} & \textbf{8.87} \\
    \bottomrule
  \end{tabular}
  }
  \vspace{0cm}
  \label{tab:m_vs_nm_fid}
  \end{center}
  \end{sc}
\end{table}

\begin{algorithm}[h!]
\small{
\caption{Sanitization algorithm}
\label{alg:sanitization}
\begin{algorithmic}[1]
\State $F \gets \emptyset$ \Comment{trained binary classifiers}
\State $NM \gets \text{deduplicated nonmembers}$
\State $M \gets \text{global set of members}$
\State $M_i \gets \emptyset$ \Comment{sanitized members after i-th iteration}
\State $TrainSet \gets \emptyset$ \Comment{training dataset} 
\For{$i \gets 1,\ 2,\ ...,\ n$}
\State $TrainSet \gets \emptyset$
\If{$i = 1$}
\State $TrainSet \gets$ sample of size $|NM|$ from $M$
\Else
\State $TrainSet \gets M_{i-1}$
\EndIf
\State $TrainSet \gets TrainSet \cup NM$
\State $F_i \gets \text{trained classifier on } TrainSet$
\While{$|M_i| < |NM|$}
\State $M_{tmp} \gets \text{sample from } M$
\For{$j \gets 1,\ 2,\ ...,\ i$}
\If{$F_j$ predicts member label for sample}
\State $M_{tmp} \gets M_{tmp} \setminus sample$
\EndIf
\EndFor
\State $M_i \gets M_i \cup M_{tmp}$
\EndWhile
\EndFor
\State $SM \gets M_n$ \Comment{final sanitized members set}
\end{algorithmic}
}
\end{algorithm}

\textbf{Santization algorithm}
The goal of this process is to create a member set that is as similar as possible to our deduplicated nonmembers set.

Main intuition behind the sanitization algorithm (Alg. \ref{alg:sanitization}) is to train a set of binary classifiers to label samples as members or nonmembers in an iterative fashion, and then pick only the samples from one of these sets, for which all of the models predict wrong label. In effect, at each iteration, one of these sets becomes closer to the other one in terms of the text embeddings distribution.

In general, we can use both members and nonmembers sets to perform this classifier-based filtering. In our case, we filter only the huge members set (LAION Aesthetics v2 5+ consists of 600M samples, our deduplicated nonmembers set has 42.5k samples). The main reason is that the deduplication process (see Sec. \ref{sec:deduplication}) is a great bottleneck of our system; to get duplicate candidates for all 160k nonmember candidates, we query the retrieval API for 50h.

\textbf{Results}
To obtain our final set of 40k sanitised members we apply the algorithm for three iterations. To obtain this subset we filter approximately 5M samples from LAION Aesthetics v2 5+ dataset, which takes only 2h on a single NVidia RTX 2080Ti. Using our assessment methodology we confirm its efficacy. FID score in the comparative case drops significantly compared to the starting members set. We see that the PCA components' distribution align (see Fig. \ref{fig:prompts_it_3}) and a binary classifier's accuracy is almost random.

\textbf{Images embeddings} Regarding the image embeddings for LAION-2B EN and LAION-2B Multi Translated, we expect them to have the same characteristics as they come from the same source. This assumption is confirmed by a low FID value between them, see Tab. \ref{tab:m_vs_nm_fid}. Additionally, the FID value decreases even further after the text-focused sanitization and a binary classifier’s accuracy is almost random. Therefore, additional sanitization efforts focused on image embeddings are not necessary, and the PCA components are also aligned, see Fig. \ref{fig:images_it_3}).

\section{Experiments}
\label{sec:attacks_setups}

\subsection{Threat model}
A membership inference attack is defined as follows. 
We consider an adversary $A$ that aims to infer whether a single
data point $x$ was included in the training set $D$ of a generative
model $M$.  The attacker has no knowledge about the dataset $D$ and is only able to query the model $M$. We distinguish three scenarios according to the
attacker's capabilities. 

\begin{itemize}
    \item In the \textbf{black-box} scenario, an adversary queries a generative model with a text prompt and gets a generated image. The attacker has no knowledge about the model architecture and no access to its weights. 
    \item In the \textbf{grey-box} scenario an adversary has access to the visual and text encoders of the attacked model. Thus, they are able to calculate the latent representation (embedding) of a given image and text prompt. However, the attacker still has no access to the model weights. 
    \item Finally, in the \textbf{white-box} scenario, an adversary has full access to the model, its source code and trained weights.
\end{itemize}

We start with a baseline white-box model loss threshold attack, which is based on the fact that machine learning models learn by minimising the loss in the training samples. We extend our analysis by covering metrics related to model inference. Moreover, we introduce new white-box attack methods and show that they outperform the commonly used baseline method. We also describe and evaluate a grey and a black-box scenario. The attacks are based on the intuition that generative models tend to synthesise samples similar to their training set. For all attacks, we evaluate a variant in which the losses are obtained as the average of 5 losses (5 different passes through the model, each time with a different noise) following the findings of~\cite{carlini2022membership}. We explore more attack methods in the Appendix \ref{sec:appendix_loss_attacks}.

\subsection{Threshold attack}
\label{sec:losses}

A general \textit{threshold} attack is formulated as follows. For a selected threshold $\tau$, the attack classifies the image $x$ as a member if $\mathcal{L} < \tau$. Otherwise, $x$ belongs to the nonmember set.
Commonly used \textit{threshold} attacks focus only on a model loss. We extend our analysis by  Pixel and Latent error, defined as follows: %

\textbf{Model loss} We monitor the loss of the diffusion model given by Eq. \ref{eq:training_loss} 
$\mathcal{L}(x, t, \epsilon; f_{\theta}) = \lVert \epsilon - f_{\theta}(x_t, t)\rVert_2^2$. 

\textbf{Pixel error} We define the pixel error as the reconstruction error between an original image $x$ and the generated image $x'$ defined as $\mathcal{L}(x,x') = \lVert x - x' \rVert_2^2$. 

\textbf{Latent error} This measurement is similar to the pixel error. However, it focuses on the reconstruction error between a latent representation $z$ of the original image $x$ and the latent representation $z'$ generated by the diffusion model. The error $\mathcal{L}(z,z')$ is defined as $\lVert z - z' \rVert_2^2$.

\begin{table*}[t!]
\vspace{-0.1cm}
\small
\centering
          \begin{sc}
                   \caption{\small \textbf{\textit{Threshold} membership inference attacks results on LAION-mi and POKEMON datasets.} We demonstrate the importance of evaluating membership inference attacks in a fair setting. On the POKEMON dataset some of the attacks are almost perfect, with \emph{partial denoising} reaching $\mathbf{99.5}$\% TPR@FPR=1\%, but on ours LAION-mi dataset with original SD-v1.4 we reach at most $\mathbf{2.51}$\%. Our proposed methods outperform the \emph{Baseline loss threshold} method.}
                        \vspace{0.2cm}

          \begin{tabular}{c c c c c}
          \toprule
               & & &  \multicolumn{2}{c}{\textbf{TPR@FPR=1\%. $\uparrow$}}\\
             \textbf{Scenario} & \textbf{Loss} & \textbf{Method} &\textbf{ LAION-mi} & \textbf{POKEMON}\\
            
            \midrule
                       \multirow{10}{*}{White-box} & \multirow{4}{*}{Model Loss}  &  Baseline loss thr. & 1.92\%$\pm$0.59 & 80.9\%$\pm2.27$ \\
            \cmidrule{3-5}

                     & & Reversed noising & 2.51\%$\pm$0.73 & 97.3\%$\pm0.93$ \\
                     &  &  Partial denoising &  2.31\%$\pm$0.61 & 94.5\%$\pm1.34$ \\
                   & & Reversed denoising & 2.25\%$\pm$0.64 & 91.5\%$\pm1.63$ \\

            \cmidrule{2-5}

                     & \multirow{3}{*}{Latent Error} & Reversed noising & 1.26\%$\pm$0.62 & 11.5\%$\pm1.84$ \\
                     &  & Partial denoising &  2.42\%$\pm$0.62 & $99.5$\%$\pm0.4$ \\
                  &  & Reversed denoising & 2.17\%$\pm$0.64 & 61.1\%$\pm2.74$ \\ 

            \cmidrule{2-5}

                     & \multirow{3}{*}{Pixel Error} & Reversed noising & 1.90\%$\pm$0.51 & 8.36\%$\pm1.66$ \\
                    &  &  Reversed denoising & $2.03$\%$\pm$0.55 & 12.0\%$\pm1.97$ \\ 
                    & & Partial denoising &  1.75\%$\pm$0.68 & $25.38$\%$\pm2.55$ \\
            \midrule

                    Grey-box & Latent Error & Generation from prompt &  0.93\%$\pm$0.41 & 7.15\%$\pm1.5$ \\
            \midrule
                    Black-box & Pixel Error &  Generation from prompt & 0.35\%$\pm$0.19 & 12.0\%$\pm1.9$ \\
            \bottomrule
          \end{tabular}
    
    \label{tab:pokemon_vs_laion_mi_thrs}
\end{sc}
\vspace{0.4cm}
\end{table*}

\subsection{Attack methods}
\label{sec:attack_setting}
We present a baseline and the best-performing attack methods evaluated in our experiments. Most methods are only applicable under the white-box scenario but we also examine the attacks in the grey- and black-box scenario. An exhaustive description and analysis of these different attack methods are given in Appendix \ref{sec:all_settings}. 

\textbf{Baseline loss threshold}
In \cite{carlini2023extracting_diffusion} the authors show that evaluating the model loss at timestep $100$ for the latent with applied noise scale $\alpha_t$ at $t=100$ yields the best results for the membership inference attacks based on model loss. We follow this method and evaluate the model loss at timestep $100$ for the latent noised with scale $\alpha_{100}$. 

\textbf{Methods exploiting the noise} There are many possible variants of adding or removing noise in the diffusion process before we make a final decision based on the loss. The intuition (or hopeful assumption) behind such attacks is that member samples would behave more robustly than nonmembers under noisy conditions. We explore many different settings and refer a reader to Appendix \ref{sec:all_settings} for details.

\textbf{Generation from prompt}
To perform this attack we pass only the prompt associated with the original images to the model. We do so to simulate a real-world scenario, where we have access to the model only via the API. In the black-box scenario, we calculate the Pixel error between the original image and the image generated by the model using the default method of $50$ timesteps.
In the grey-box scenario, we obtain the generated images in the same way as in the black-box scenario, but then we calculate the latent representation of the original and generated images using the visual encoder of the attacked model and then calculate the latent error between them.

\subsection{Targeted datasets}
\textbf{LAION-mi} 
\label{par:eval_setup} In our paper we use our LAION-mi dataset proposed in Section \ref{sec:laion_mi}. For each attack, we use the same subset of $5000$ member samples and $5000$ nonmembers samples, further referred to as the attack set. 
We find that different training and evaluation set splits can produce significantly different results, some almost $10$ times better than others (see Appendix \ref{sec:appendix_d}). Following these findings, we evaluate our attacks on 100 random subsets (evaluation sets) of $500$ members and $500$ nonmember samples drawn from the attack set and then report the mean and the standard deviation of the performance. 

\textbf{POKEMON} POKEMON dataset~\cite{pinkney2022pokemon} is a Text2Image dataset. We use it to first finetune the original StableDiffusion v1.4 model using a subset of $633$ samples (members), leaving the remaining $200$ samples as nonmembers. We evaluate our attacks on $1000$ subsets obtained from random $200$ member and all nonmember samples. We also conduct an analysis of the influence of overfitting on the attack performance in Appendix \ref{sec:appendix_e}.

\section{Results}

We evaluate the attacks for two setups. First, we attack the Stable Diffusion v1.4 model, which we do not modify in any way. We draw data samples of members and nonmembers from our LAION-mi dataset. Then, we evaluate the effectiveness of the attacks on the same model, which is finetuned on the POKEMON dataset~\cite{pinkney2022pokemon}. Here, we use test and training data splits ($633$/$200$ samples) as member and nonmember sets.

\textbf{Metric} A metric to evaluate the membership attack is true-positive rate (TPR) calculated at a low false-positive rate (e.g. FPR=1\%). For privacy-related problems, it is a much better metric than common aggregate metrics, such as accuracy or AUC~\cite{carlini2022membership}.

\textbf{Discussion}
In Table \ref{tab:pokemon_vs_laion_mi_thrs} we observe a severe discrepancy in the effectiveness of the attacks achieved for the LAION-mi dataset and fine-tuning on POKEMON. In the second case, two white-box attacks (\emph{\textit{reversed noising}} and \emph{\textit{partial denoising}}) achieve a very high TPR. However, when the attacks are applied against our LAION-mi dataset, we observe a huge drop in performance. Here we clearly see the effects of the first pitfall from Sec. \ref{sec:attack_challenges}, namely the fine-tuned model overfits, and the membership inference task becomes trivial. We explore this topic further in Appendix \ref{sec:appendix_e}. The obtained results demonstrate that evaluation on a small dataset (used for finetuning the model) is misleading and a more careful setup, such as our proposal, is required. 

Moreover, the results show the limited performance of loss-based attacks in the black- and grey-box scenarios, even for the simple POKEMON setting. This highlights an important issue, as many image-generation services work as a black-box API. As mentioned in \ref{sec:loss_based} this trend is unlikely to change in the future. 

In theory, identifying training samples in black-box scenarios can be approached by extracting training samples from the model, as in \cite{carlini2023extracting_diffusion}. However, this approach requires the generation of hundreds of images per text prompt, which is computationally expensive. This approach is also limited to identifying training samples that have been memorised by the model. For those reasons, such methods are not well suited for identifying the membership of a sample. For the white-box scenario, the state-of-the-art approach is a method based on the shadow models. However, as stated in \ref{sec:shadow_models}, this strategy is too costly for large diffusion models such as Stable Diffusion. We further discuss the applicability of shadow models in Appendix \ref{sec:appendix_shadow}.

\section{Conclusion}

We showed that evaluation of membership inference attacks with the model finetuning approach may lead to false conclusions. As an alternative, we proposed a new carefully crafted dataset, which mitigates the main limitation of the original LAION dataset, which is a lack of a test set. Having the proposed dataset and reliable set of nonmembers, we evaluated several membership inference attacks and obtained results, which contradict previous findings.
 
Our dataset could help the community evaluate attacks on large generative diffusion models such as Stable Diffusion in a more rigorous and fair setting. A clear picture of how successful the membership attacks are is essential for a sound policy on matters such as data ownership and privacy. We argue that for large diffusion models, where shadow models are prohibitively expensive, membership inference remains a very challenging task.

{
\bibliographystyle{ieee_fullname.bst}
\bibliography{main.bib}
}

\newpage 
\clearpage
\newpage 
\appendix

\section{Appendix: Broader impact}
\label{sec:appendix_broader_impact}
In this work, we treat the membership inference attacks as a potential tool for privacy protection. By applying these attacks to a model, we can highlight the instances where images have been used in training without the appropriate consent. As more individuals and organizations become aware of this possibility, we might see a push towards more stringent data usage policies and ethical guidelines for machine learning practices. The potential of these attacks to expose privacy violations could serve as a catalyst for a broader dialogue on privacy rights and data ownership in the digital age.

However, these benefits come with a significant caveat. The same mechanism that can be used to protect user privacy can also be used maliciously. If a membership inference attack is successful and achieves high accuracy, it could potentially lead to personal data leakage. Thus, the dual implications of membership inference attacks for diffusion models offer both a warning and a beacon. They highlight the need for robust privacy protection measures while simultaneously alerting us to the potential risks of personal data leakage. 

\section{Appendix: Limitations}
\label{sec:appendix_limitations}
LAION-5B dataset, the source of data used both to train the Stable Diffusion model and to create LAION-mi dataset does not contain the images, just URLs to them. Therefore, to prepare a training set, one needs to first download the images using the URLs, and then train the model. Since the content, that a URL points to is out of our control, it might be a subject of edition or deletion, which we are unable to handle. Because of that, we cannot guarantee with $100\%$ certainty that every member in LAION-mi is a real member of the Stable Diffusion model's train set. 

In fact, during our experiments, we find out that approximately $10\%$ of links are dead, i.e. we cannot download images from them. This limitation is inherent to the LAION-5B dataset, and therefore cannot be alleviated. Additionally, a URL to an image from LAION-mi has to be alive to use it for the attack evaluation. We argue that it is unlikely that a URL from the LAION-mi members subset that was dead during training of the Stable Diffusion is now alive, however, we cannot rule out such scenario. Fortunately, since it affects only the members set, it makes the membership inference task harder, in effect saving us from the pitfalls described in Section \ref{sec:attack_challenges}.

\section{Appendix: Stable Diffusion-v1.4 training}
\label{sec:appendix_sd14}
\textbf{Datasets} Datasets used to train this model are as follows
\begin{itemize}
    \item LAION-2B EN: a subset of the LAION-5B\cite{schuhmann2022laion5b} with images' descriptions in English.
    \item LAION-Aesthetics V2 5+\cite{laion-aesthetics}: a subset of the LAION-2B EN dataset, in which each image is scored using the LAION-Aesthetics\_Predictor V2\cite{aesthetic_predictor}, and only images with the Aesthetic Score above 5 are a part of it.
\end{itemize}

\textbf{Stable Diffusion-v1.4} This version of the Stable Diffusion is first trained for 431k steps using samples from LAION-2B EN, then fine-tuned for 515k steps using LAION-Aesthetics V2 5+, and then fine-tuned for 225k steps on LAION-Aesthetics V2 5+ again \cite{sd-v14-model-card}.

\section{Appendix: Loss attacks}
\label{sec:appendix_loss_attacks}
In this section, we discuss different approaches to perform the diffusion denoising process in the white-box scenario. In each of the following setups, at every timestep, we collect the Model loss, Latent error, and Pixel error described in Section \ref{sec:losses}. These values are calculated based on the output of the UNet SD-v1.4 backbone using the methods described below.

\subsection{All attack methods}
\label{sec:all_settings}
Here we describe all the methods used in our experiments. For all methods, the classifier-free guidance is applied with scale $7.5$ unless otherwise stated.

\begin{enumerate}
    \item \textbf{Partial denoising} Following findings in \cite{carlini2023extracting_diffusion} in this method, we start our denoising process at the timestep $300$ for the steps $26$ up to the timestep $50$. Latent image representation is noised once, with scale $\alpha_{300}$, and then for each step calculated from the UNet noise prediction.

    \item \textbf{Partial denoising non-iterative} is similar to the \textit{Partial denoising} method, but at each timestep we pass a newly noised latent representation of the image to the UNet, instead of the output of the previous timestep's inference through UNet. The main intuition behind this is to see whether the information about membership gets lost or accumulates during the iterative denoising process.
    \item \textbf{Partial denoising latent shift} follows the \textit{Partial denoising} attack method, but at the middle timestep $170$ we shift the latent representation of the image by a scaled random noise. The intuition behind this method is that the member samples should return to the correct trajectory after the shift better than the nonmember samples enabling us to better identify the member samples.
    \item \textbf{Partial denoising text shift} is similar to the \textit{Partial denoising} latent shift, but instead of shifting the latent representation of the image, we shift the text embedding by a random noise $N(0, I)\times0.1$ before passing it to the UNet. The intuition here is the same as in the partial denoising latent shift method, but we want to see if the text embedding is more important than the latent representation of the image.
    \item \textbf{Partial denoising bigger text shift} is identical to the \textit{Partial denoising text shift}, but the noise added to the text embedding is scaled by $0.5$ instead of $0.1$. We want to test if adding more noise to the text embedding benefits the performance of the attacks.
    \item \textbf{Partial denoising wrong start} In this method we follow the \textit{Partial denoising}, but the starting latent representation of the image is noised using $\alpha_t$ from the wrong timestep, $t=100$ instead of $t=300$. We want to test whether we can extract more information about the membership if we apply lower noise to the latent representations at the beginning of the process than UNet expects.
    \item \textbf{Full denoising start 100} is similar to the \textit{Partial denoising wrong start} method, but we perform an almost full denoising process, starting on timestep and $900$ ending at timestep $0$, denoising every $100$ and starting from the latent noised with $\alpha_{100}$ noise scale.
    \item \textbf{Full denoising start 100 text shift} is close  to the \textit{Full denoising start 100}, but we shift the text embedding by a random noise $N(0, \mathbf{I})*0.1$ before passing it to the UNet.
    \item \textbf{Full denoising start 50} is similar to the Full denoising start 100, but the starting latent is noised using $\alpha_{50}$ instead of $\alpha_{100}$.
    \item \textbf{Full denoising start 300} is like the Full denoising start 100, but the starting latent is noised using $\alpha_{300}$ instead of $\alpha_{100}$.
    \item \textbf{Short denoising start 300}  In this method, we perform a short denoising process, starting from timestep $200$ ending at timestep $0$ every $100$ steps instead of the full one. The latent representation of the image is noised using $\alpha_{300}$.
    \item \textbf{Reversed noising}
We perform $10$ denoising steps for timesteps from $900$ to $0$. The image latent representation we get from the VAE encoder is noised using noise scales $\alpha_{t}$ in reverse order, e.g. at timestep $800$ we pass the latent noised with $\alpha_{100}$ to the UNet.
Additionally, during inference, we use classifier-free guidance on the guidance scale $100$. The text embedding passed to UNet remains unchanged. The intuition behind the attack is that member samples will behave more robustly than nonmember samples under such conditions. 
    \item \textbf{Full denoising start 300 no cfg} is similar to the \textit{Full denoising start 300}, but we do not use the classifier-free guidance. We want to see if the classifier-free guidance is beneficial for the attacks performance.
    \item \textbf{Full denoising start 100 non-iterative}. This method is identical with the \textit{Full denoising start 100}, but at each timestep we pass newly noised latent representation with scale $\alpha_{100}$ of the image to the UNet, instead of the output of the previous timestep.
    \item \textbf{Reversed noising regular cfg} resembles the \textit{Full denoising start 100 non-iterative} attack method. The latent representations we pass to the UNet are noised using $\alpha_t$ from the timesteps in the reversed order, i.e. from $0$ to  $900$, so the first latent passed to the UNet is noised using $\alpha_{0}$, while UNet receives timestep $900$ as input. The classifier-free guidance is applied with scale $7.5$.
    \item \textbf{Reversed denoising}
In this method, we reverse the order of timesteps at which we perform denoising using UNet. We start from the timestep $0$ and then go up to the timestep $900$, for $10$ steps in total. Similarly as in the baseline loss threshold method, we apply noise to the latent image representation once, but this time using noise scale $\alpha_{100}$. 
At each timestep we measure all losses described in Section \ref{sec:losses} and use them to evaluate the attack. The intuition here is similar as in the \textit{Reversed noising} method, but here we test if the iterative nature of the diffusion denoising process will magnify this effect.

\end{enumerate}

\subsection{Classifier attack}
\label{sec:classifiers_appendix}

\begin{table}[b!]

\caption{\small \textbf{Hyperparameters used for the classifiers:} Logistic Regression (LR), Decision Tree Classifier (DTC) and Random Forest Classifier (RFC).}
\label{tab:hyperparameters_classifiers}
    \centering
    {
        \begin{tabular}{cccc}
            \toprule
            Classifier               & Hyperparameter      & Values                  \\
            \midrule
            LR     & C                   & [0.01, 0.1, 1, 10, 100] \\
            DTC & max\_depth          & [2, 8, 16]              \\
                                     & min\_samples\_split & [2, 8, 16]              \\
            RFC & n\_estimators       & [10, 100, 1000]         \\
                                     & max\_depth          & [2, 8, 16]              \\
            \bottomrule
        \end{tabular}
    }
\end{table}

\begin{table*}[b!]
    
    \caption{ \textbf{\textit{Threshold} attack results for each method.} We see that for each attack method Model loss gives out better performance of \textit{threshold} attacks compared to Latent or Pixel error. It is also more robust for the denoising procedure modification than Latent Error, dropping to at most $2\%$, while Latent error performance can drop to $1.1\%$. Pixel error seems also more stable than Latent Error, but we cannot achieve as high results as for the Model loss or Latent error. The high discrepancy between different attack methods points that we can indeed influence the amount of membership information extracted during influence, with some methods performing better on Model loss and some on other losses, Pixel and Latent error.}
    \label{tab:settings_results_thrs}
    \begin{sc}
        \centering
            {
            \begin{tabular}{cccc}
    \toprule
    Loss                                & Model Loss      & Latent Error    & Pixel Error     \\
    method                             &                 &                 &                 \\
    \midrule
    Partial denoising                   & 2.3\%$\pm$0.61  & $\mathbf{2.4}$\%$\pm$0.62  & 1.7\%$\pm$0.68  \\
    Partial denoising non-iterative     & 2.3\%$\pm$0.68  & 1.1\%$\pm$0.46  & 1.39\%$\pm$0.59 \\
    Partial denoising latent shift      & 2.3\%$\pm$0.62  & 2.24\%$\pm$0.9  & 1.6\%$\pm$0.62  \\
    Partial denoising text shift        & 2.2\%$\pm$0.57  & 2.28\%$\pm$0.57 & 1.7\%$\pm$0.64  \\
    Partial denoising bigger text shift & 2.3\%$\pm$0.62  & 2.22\%$\pm$0.75 & 1.74\%$\pm$0.6  \\
    Partial denoising wrong start       & 2.2\%$\pm$0.69  & 2.13\%$\pm$0.85 & 1.99\%$\pm$0.56 \\
    Full denoising start 100            & 2.08\%$\pm$0.64 & 1.1\%$\pm$0.41  & 1.84\%$\pm$0.6  \\
    Full denoising start 100 text shift & 2.01\%$\pm$0.6  & 1.1\%$\pm$0.43  & 1.8\%$\pm$0.6   \\
    Full denoising start 50             & 2.0\%$\pm$0.6   & 1.15\%$\pm$0.38 & 1.95\%$\pm$0.55 \\
    Full denoising start 300            & 1.99\%$\pm$0.61 & 1.34\%$\pm$0.49 & 1.72\%$\pm$0.64 \\
    Short denoising start 300           & 2.32\%$\pm$0.67 & 2.06\%$\pm$0.72 & 1.57\%$\pm$0.67 \\
    Reversed denoising                  & 2.25\%$\pm$0.64 & 2.17\%$\pm$0.64 & $\mathbf{2.03}$\%$\pm$0.55 \\
    Full denoising 300 no cfg           & 2.07\%$\pm$0.62 & 1.45\%$\pm$0.52 & 1.7\%$\pm$0.6   \\
    Full denoising 100 non-iterative    & 2.2\%$\pm$0.55  & 2.18\%$\pm$0.75 & 1.91\%$\pm$0.53 \\
    Reversed noising regular cfg        & $\mathbf{2.5}$\%$\pm$0.74  & 2.03\%$\pm$0.6  & 1.92\%$\pm$0.5  \\
    Reversed noising                    & $\mathbf{2.51}$\%$\pm$0.73 & 1.26\%$\pm$0.52 & 1.9\%$\pm$0.51  \\
    \bottomrule
\end{tabular}
}

    \end{sc}
\end{table*}

In addition to the \textit{threshold} attack type described in \ref{sec:losses}
we also introduce the \textit{classifier} attack type. \textit{Classifier} attack builds on top of the \textit{threshold} attack. It uses a binary classifier $C$ to predict the membership of an image $x$ based on the losses collected during inference through the diffusion model. The classifier is trained on a set of images with known membership and returns a probability of membership for a given image. We then perform a \textit{threshold} attack on the classifier's output.

We train four model classes: Logistic Regression (LR), Decision Tree Classifier (DTC), Random Forest Classifier (RFC), and Neural Network binary classifier (NN). The classifiers are trained in the binary classification task, with the member samples as positive examples and the nonmember samples as negative examples. The input for every classifier consists of the vector of the loss values we collect at every timestep. For the Logistic Regression (LR), Decision Tree Classifier (DTC) and Random Forest Classifier (RFC) we perform Grid Search with k-Cross Validation, $k=5$ for each fit. The hyperparameters used for the classifiers are described in Table \ref{tab:hyperparameters_classifiers}. Note that we perform the Grid Search for every fit separately, therefore we do not report the best parameters in our table, since they turn out to be different for different fits.

For the Neural Network classifier (NN) we use the following architecture: 3 fully connected layers, with the input size of $3*timesteps$ (since for different methods we have a different amount of data), hidden size of $10$ and a single output for the binary classification. We use ReLU activation function for the hidden layers and Sigmoid for the output layer. We use Adam optimizer with a learning rate $0.001$ and binary cross entropy loss. We train the classifier for $100$ epochs with batch size $32$ and early stopping. We use the best model from the early stopping for the evaluation.

Following \ref{par:eval_setup} we fit our models $100$ times on $100$ different training sets sampled from the whole attack set and evaluate them on the remaining evaluation sets. We report the mean and standard deviation of the metrics for the $100$ fits. The results are presented in Table \ref{tab:settings_results_thrs}.

\subsection{Results}

\begin{table*}[ht!]
    
    \caption{ \small\textbf{\textit{Classifier attack} results for each method.} Same as in Tab. \ref{tab:settings_results_thrs} we observe that for different attack methods the same model classes achieve visibly different performance. This again confirms that we are able to obtain more information about the membership of samples by influencing the whole denoising process of the large diffusion model. Unsurprisingly, we are able to outperform the simple \textit{threshold} attacks, because the membership information ends up spread out on different timesteps and losses. Using \textit{classifier} attack we can combine the information from the whole inference procedure and make better predictions. Surprisingly, it seems to be harder to perform such attacks in the TPR@FPR=1\% regime. For almost all methods and model classes we are not able to outperform the simple \textit{threshold} attacks, especially on the Model loss. For almost all methods the Random Forest Classifier model outperforms all other model types, with the exception of \textit{Reversed noising regular cfg} method data fitted using Logistic Regression. Decision Tree Classifier fails to be better than a random guess for almost all methods, and the Neural Network classifier also fails to deliver satisfying results.}
    \label{tab:settings_results_clf}
        \centering
    
    \begin{sc}
\begin{tabular}{ccccc}
\toprule
Classifier class &               LR &              DTC &              RFC &               NN \\
method                             &                  &                  &                  &                  \\
\midrule
Partial denoising                   &  1.83\%$\pm$0.87 &  0.67\%$\pm$1.09 &  2.27\%$\pm$0.76 &  1.50\%$\pm$0.92 \\
Partial denoising non-iterative     &  2.24\%$\pm$0.86 &  $\mathbf{1.10}$\%$\pm$1.12 &  2.35\%$\pm$0.88 &  1.52\%$\pm$1.03 \\
Partial denoising latent shift      &  1.94\%$\pm$0.92 &  0.86\%$\pm$0.94 &  2.35\%$\pm$0.86 &  1.40\%$\pm$0.91 \\
Partial denoising text shift        &  1.71\%$\pm$0.87 &  0.49\%$\pm$0.84 &  2.43\%$\pm$0.93 &  1.40\%$\pm$1.00 \\
Partial denoising bigger text shift &  2.14\%$\pm$1.01 &  0.80\%$\pm$1.13 &  2.30\%$\pm$0.73 &  1.49\%$\pm$0.90 \\
Partial denoising wrong start       &  1.50\%$\pm$0.70 &  0.74\%$\pm$1.08 &  2.26\%$\pm$0.87 &  1.51\%$\pm$0.92 \\
Full denoising start 100            &  1.09\%$\pm$0.56 &  0.54\%$\pm$0.74 &  1.89\%$\pm$0.85 &  1.30\%$\pm$0.76 \\
Full denoising start 100 text shift &  1.10\%$\pm$0.62 &  0.62\%$\pm$0.76 &  1.88\%$\pm$0.80 &  1.24\%$\pm$0.69 \\
Full denoising start 50             &  1.16\%$\pm$0.68 &  0.70\%$\pm$0.68 &  1.89\%$\pm$0.78 &  1.31\%$\pm$0.78 \\
Full denoising start 300            &  1.21\%$\pm$0.58 &  0.42\%$\pm$0.85 &  1.87\%$\pm$0.87 &  1.39\%$\pm$0.80 \\
Short denoising start 300           &  1.94\%$\pm$0.91 &  0.54\%$\pm$0.83 &  2.31\%$\pm$0.91 &  1.49\%$\pm$0.85 \\
Reversed denoising                  &  1.96\%$\pm$0.96 &  0.80\%$\pm$1.17 &  2.37\%$\pm$0.98 &  1.26\%$\pm$0.78 \\
Full denoising 300 no cfg           &  1.31\%$\pm$0.73 &  0.31\%$\pm$0.65 &  1.78\%$\pm$0.78 &  1.37\%$\pm$0.82 \\
Full denoising 100 non-iterative    &  1.60\%$\pm$0.77 &  0.68\%$\pm$1.05 &  2.19\%$\pm$0.85 &  1.60\%$\pm$0.73 \\
Reversed noising regular cfg        &  $\mathbf{2.41}$\%$\pm$1.09 &  0.47\%$\pm$1.03 &  2.17\%$\pm$0.84 &  1.40\%$\pm$0.82 \\
Reversed noising                    &  1.98\%$\pm$0.97 &  0.41\%$\pm$1.02 &  $\mathbf{2.75}$\%$\pm$1.03 &  $\mathbf{1.62}$\%$\pm$0.86 \\
\bottomrule
\end{tabular}
    \end{sc}
\vspace{0.1cm}
    
\end{table*}

\begin{figure*}[h!]
    \centering
    \includegraphics[width=0.6\textwidth]{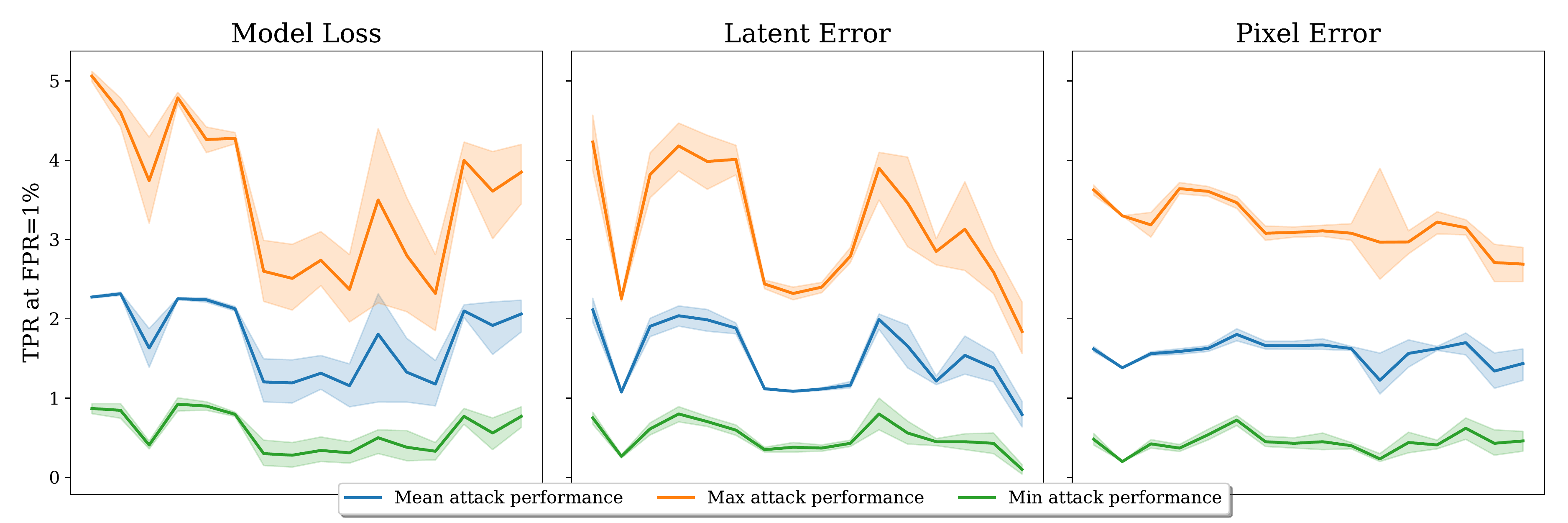}
    \caption{\textbf{Min, mean, max TPR@FPR=1\% for different methods and losses, \textit{threshold} attack type.} Solid line indicates the mean value of the TPR@FPR=1\% metric from 100 experiments, and the shaded area is the 95\% confidence interval.}
    \label{fig:randomization_importance}
\end{figure*}

\begin{figure*}[t!]
    \centering
    \includegraphics[width=0.8\textwidth]{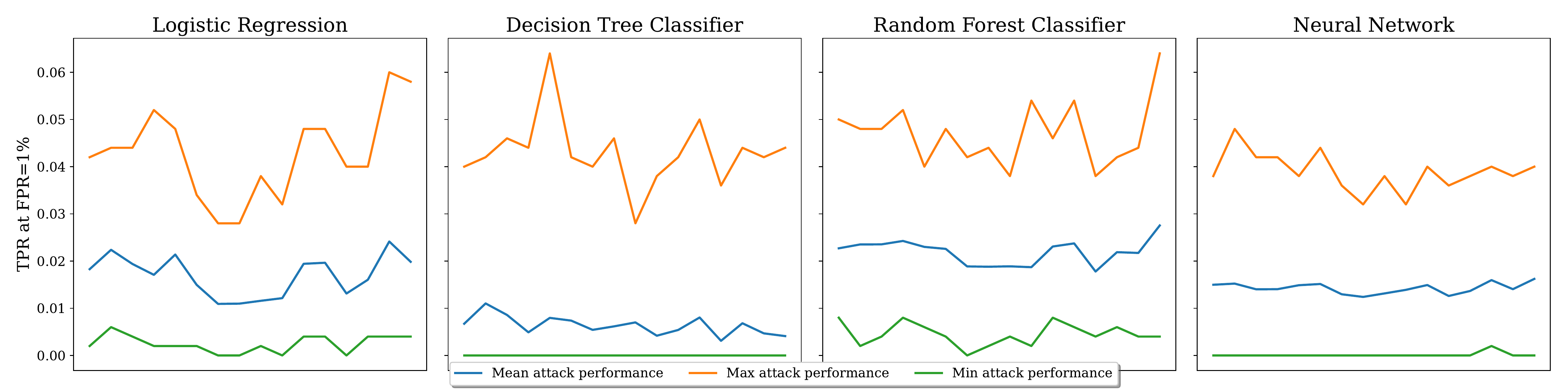}
    \caption{\textbf{Min, mean, max TPR@FPR=1\% for different methods, classifier classes and losses, \textit{classifier attack} type.}}
    \label{fig:randomization_importance_clf}
\end{figure*}

\begin{figure*}[h!]
    \centering
    \includegraphics[width=0.79\textwidth]{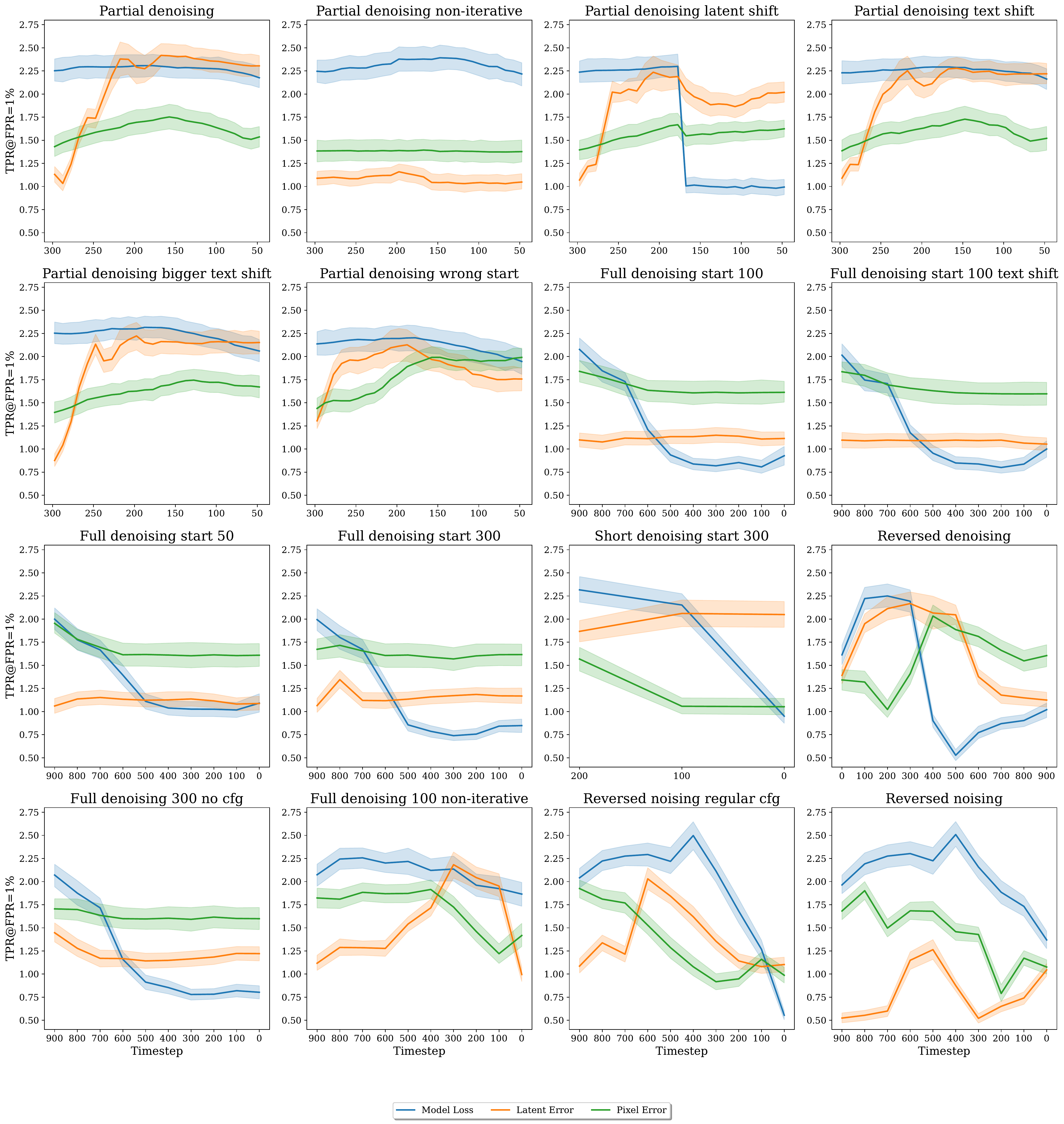}
    \caption{\small{ \textbf{Attack performances on different losses in different timesteps for each attack method for the \textit{threshold} attack type.} The solid line indicates the mean value of the TPR@FPR=1\% metric from 100 experiments, and the shaded area is the 95\% confidence interval. We see how important is to collect all three losses on each timestep, with the biggest difference visible for the \textit{Reversed noising} method, with the worst result around $0.5\%$ and the best around $2.5\%$ for Latent error at timestep $0$ and Model loss at timestep $500$. We derive the following insights from these plots. Applying the denoising process iteratively instead of passing the newly noised latent to the UNet greatly benefits the Latent and Pixel error \textit{threshold} attacks, while slightly reducing the effectiveness of \textit{threshold} attacks on the Model loss. Shifting the latent by a random noise mid-inference affects the results negatively, with the biggest drop in performance for the Model loss. Shifting text embedding by a random noise for different noise scales ($0.1$ and $0.5$) does not make any significant difference in terms of the final results, generally being slightly harmful to the performance of the \textit{threshold} attack. The mismatch between the actual and latent noise timesteps is by far the most impactful modification we implemented. We see improvements on the Pixel error \textit{threshold} attacks for \textit{Partial denoising wrong start} method, where we start from the latent noised with the noise scale of $\alpha_t=100$ and denoising from the timestep $300$ to $50$. However, for the \textit{Full denoising} attack methods we see that different noise scales applied to the starting latent representations do not change the performance. Applying the classifier-free guidance seems to be insignificant for the final results (see \textit{Full denoising 300} vs \textit{Full denoising 300 no cfg}), but increasing the guidance scale (from the default $7.5$ to $100$) reduces the effectiveness of the \textit{threshold} attacks on Pixel and Latent error. The best results are obtained when we reverse the order of noise scales used to apply noising on the latent representation of the input image, while the input timesteps we pass to UNet remain in the normal order. We also see in Tab. \ref{tab:settings_results_clf} that these methods allow the \textit{classifier} attacks to achieve the best results, suggesting that this approach extracts the most data about the membership from the model.}}
    \label{fig:full_settings_overview}
\end{figure*}

Following our evaluation method described in Section \ref{par:eval_setup} we report our results for each method in Table \ref{tab:settings_results_thrs} for \textit{threshold} attacks, and in Table \ref{tab:settings_results_clf} for \textit{classifier} attacks.

We conclude that extraction of the membership information from the attacked model can be improved by modifying the diffusion denoising process by altering the timesteps, latent representations, and text embeddings. We also see that the information about membership can be better extracted by using \textit{classifier} attack, but for most methods, a simple \textit{threshold} attack outperforms the \textit{classifier} attack.

We also observe that different methods achieve visibly different performance on different timesteps, which points out that the timing of loss measurement is also really important when performing the \textit{threshold} attack, see Figure \ref{fig:full_settings_overview}.

\section{Appendix: Experiments randomization}
\label{sec:appendix_d}

Here we highlight the need to perform the randomization described in Section \ref{par:eval_setup}. For each of the methods described in Appendix \ref{sec:appendix_loss_attacks} we show the differences between the best, mean and worst results for each attack from the separate runs. We also visualise these differences for the classification attack type. We can observe a mismatch between mean and best results for all of the attacks, some of them being even three times worse than the best ones (\textit{Partial denoising} method). When the available size of the nonmembers set is relatively small, the randomization is crucial to obtain reliable results, especially when proposing attacks based on more sophisticated methods that require training a classifier, which is the case in the \textit{classifier} attack type. In this case, we suggest splitting the whole attack set randomly in the way described in Section \ref{par:eval_setup} into training and evaluation sets, then training the classifier, and repeating the whole procedure for at least $100$ times. The reported results should be the mean of the results for each attack from each run. In this way, we can mitigate the influence of the potential outliers and obtain more reliable results.

Visualisation of the mismatch between the best, mean, and worst results can be found in Figure \ref{fig:randomization_importance} for the \textit{threshold} attacks and Figure \ref{fig:randomization_importance_clf} for the \textit{classifier} attacks.

\section{Appendix: Overfitting impact on membership inference attacks}
\label{sec:appendix_e}
\begin{figure}[t!]
\centering
\begin{subfigure}{0.45\textwidth}
        \includegraphics[width=\textwidth]{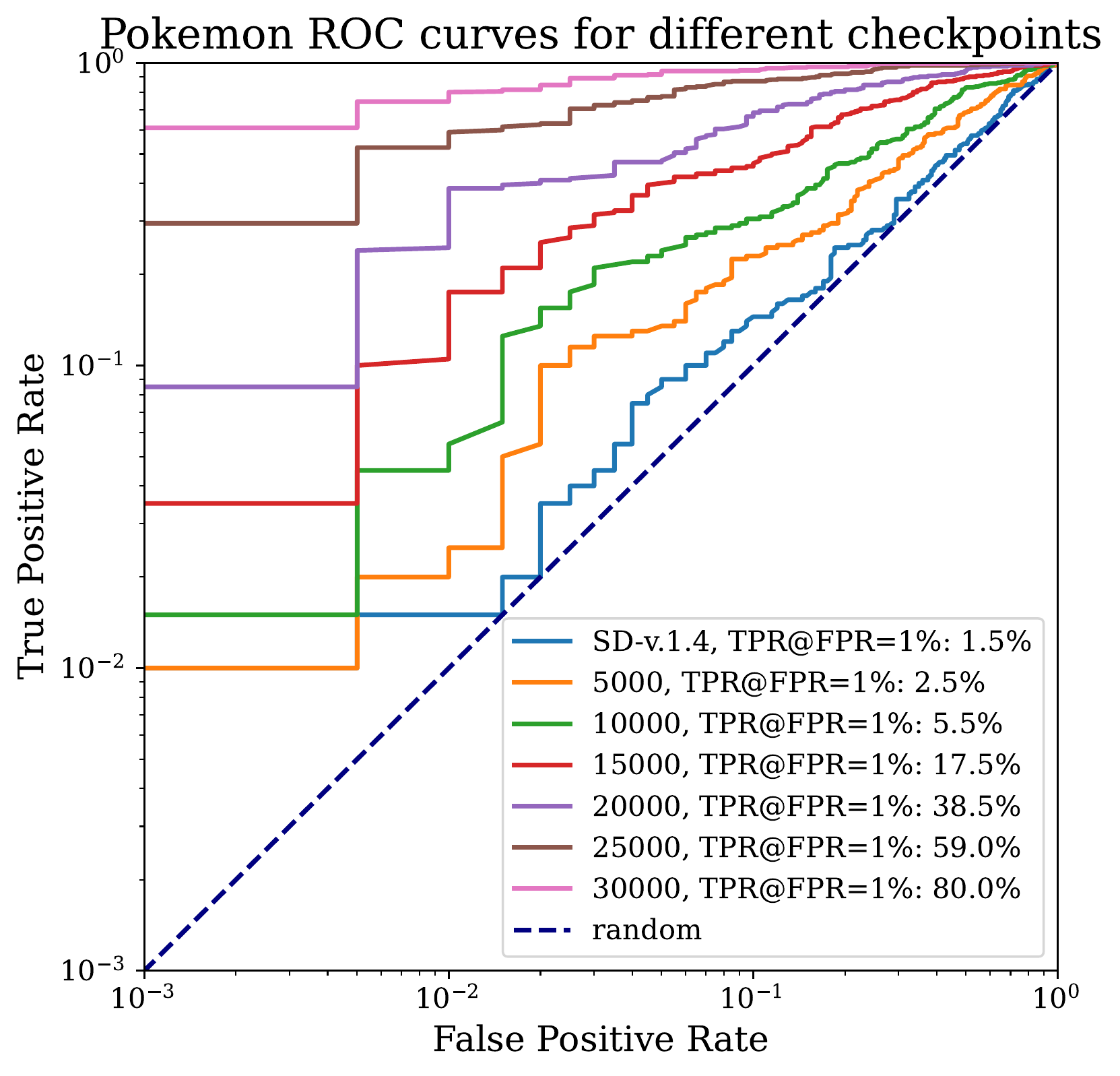}
        \caption{\textbf{ROC curves and TPR@FPR=1\% results for different checkpoints.}}
        \label{fig:pokemons_roc}
\end{subfigure}

\begin{subfigure}{0.45\textwidth}
        \includegraphics[width=\textwidth]{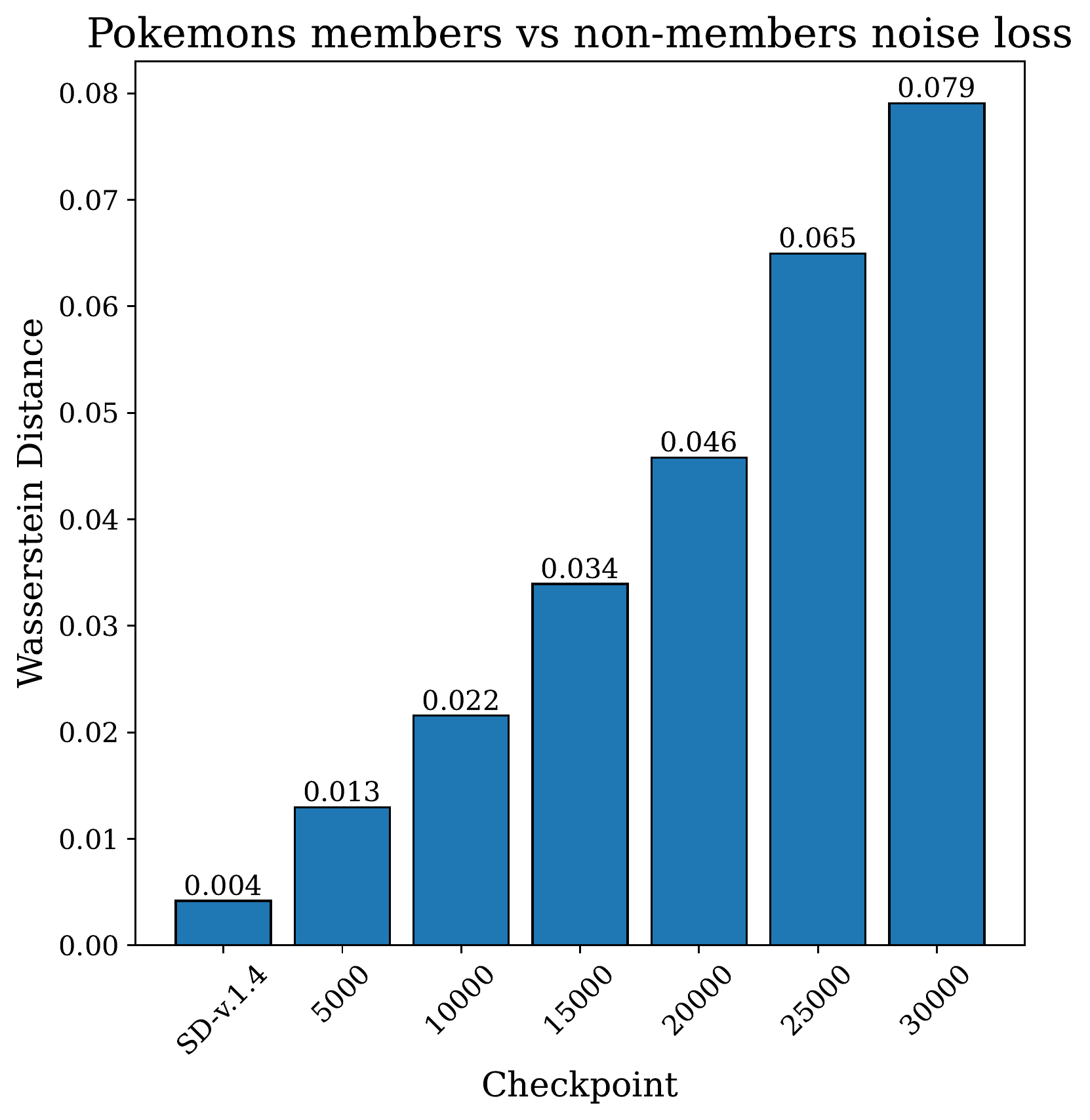}
        \caption{\small \textbf{Wasserstein Distance for different checkpoints.}}
        \label{fig:ws_d_pokemons}
\end{subfigure}
    \caption{\small Figure \ref{fig:pokemons_roc} shows how overfitting correlates with the attacks performance. Figure \ref{fig:ws_d_pokemons} shows the Wasserstein Distance calculated between the losses of members and nonmembers sets for each checkpoint.}
    \label{fig:overfitting_generalization}
\end{figure}

In this section, we present the results of our experiments on overfitting on the POKEMON dataset from Section \ref{par:eval_setup}. We finetune the original StableDiffusion-v1.4 using the default method from \cite{pokemon_finetuning} for $30000$ train steps with the learning rate $0.00001$ and gradient accumulation of $4$. Every $5000$ steps we save the partially finetuned model.

The attack method we use to perform for this section is the Baseline loss threshold method, described in Section \ref{sec:attack_setting}. The ROC curves and TPR@FPR=1\% results for different checkpoints can be found in Figure \ref{fig:pokemons_roc}. We can observe that the model is overfitting to the training set, as the TPR@FPR=1\% results are significantly better for the checkpoints from the later stages of the training. This is in line with the findings in \cite{carlini2022membership}, which also show that model loss based membership inference attacks give better results when the model is overfitted. This issue seems to be not addressed in some of the previous contributions, e.g. in \cite{duan2023diffusion} authors claim really good performance of their attacks on the finetuned version of the SD-v1.4 model on the same dataset.

\section{Appendix: Fine-tuned Shadow models for Large Diffusion Models}
\label{sec:appendix_shadow}
Training a new Stable Diffusion model from scratch multiple times is in practice infeasible. Thus, it is impossible to directly apply the existing shadow model attack in this case. To overcome this limitation we focus our analysis on model fine-tuning, rather than training from the start. Motivated by the state-of-the-art results achieved by shadow model membership inference, we draw inspiration for our approach from the offline \textit{LIRA} attack introduced in \cite{carlini2022membership}. Intuitively, we aim to answer the question "\textit{is there a difference between finetuning the model on member and nonmember samples?}". To overcome the memory requirements for storing multiple versions of the Stable Diffusion models we apply fine-tuning with \textit{LoRA}~\cite{hu2021lora}. 

\begin{figure*}[t!]
    \begin{subfigure}{0.49\textwidth}
        \centering
        \includegraphics[width=\textwidth]{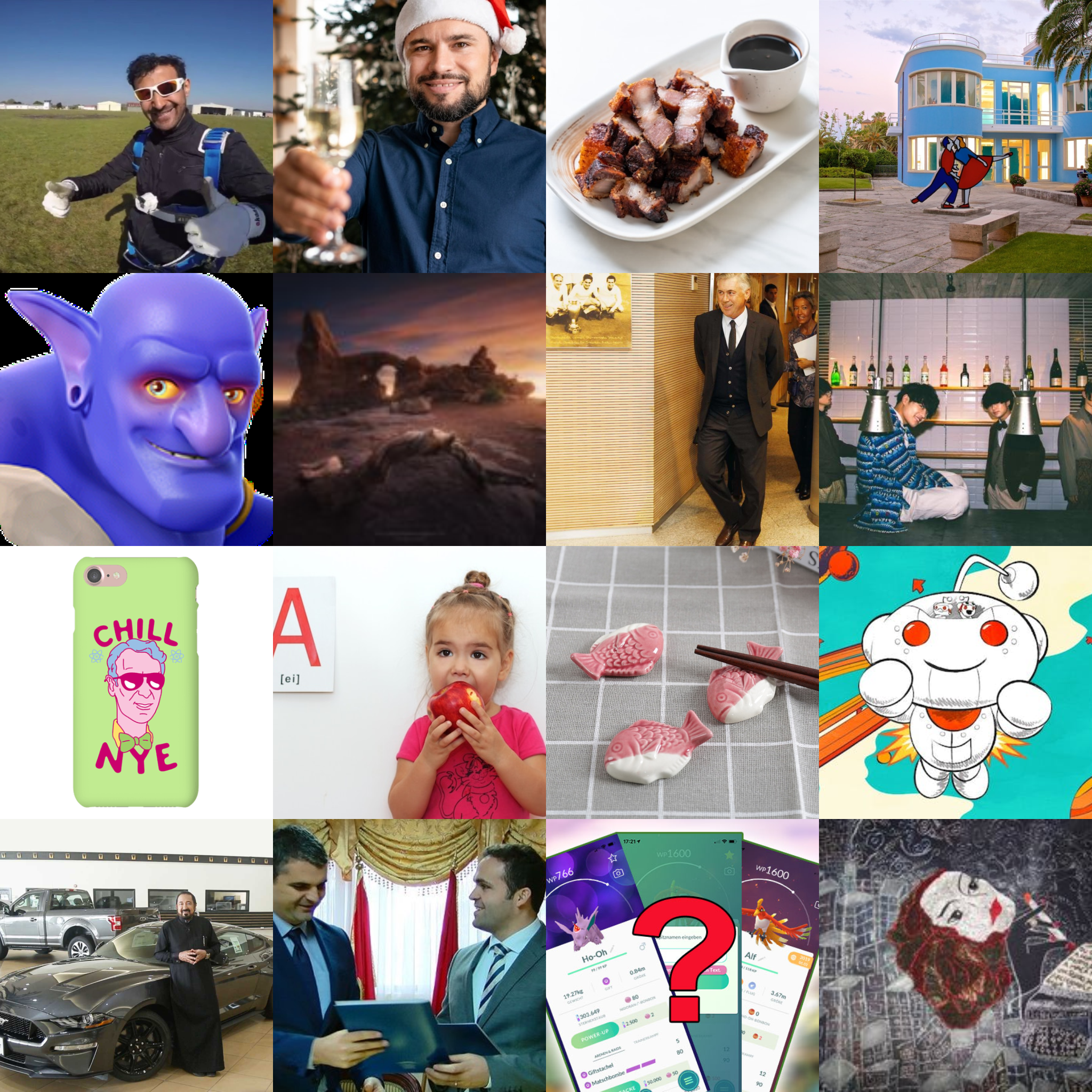}
        \caption{LAION-mi nonmember samples.}
        \label{fig:laion_mi_nm}
    \end{subfigure}
    \begin{subfigure}{0.49\textwidth}
        \centering 
        \includegraphics[width=\textwidth]{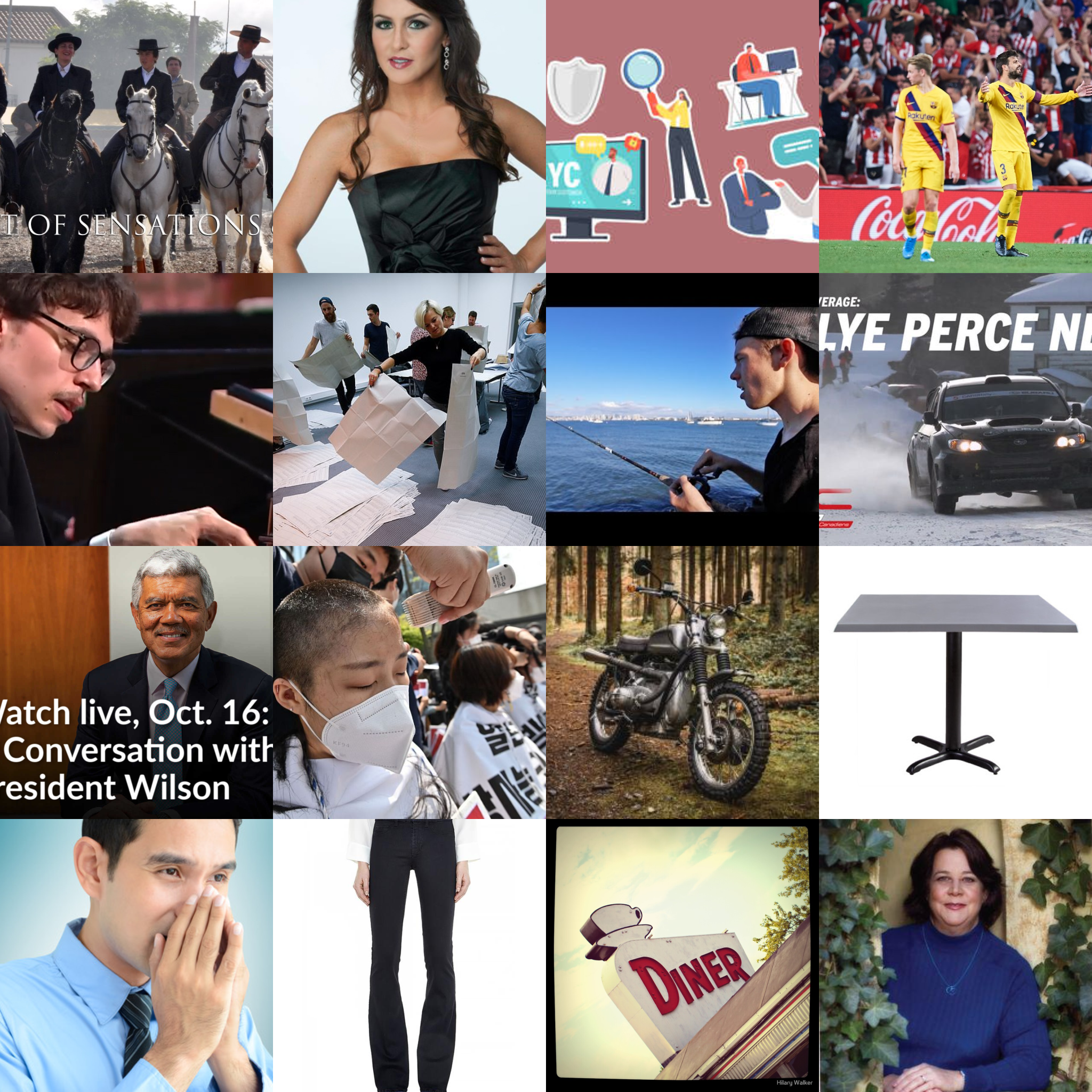}
        \caption{LAION-mi member samples.}
        \label{fig:laion_mi_m}
    \end{subfigure}
    \caption{Visualization of the random subset of LAION-mi dataset}
    \label{fig:laion_mi_imgs}
\end{figure*}

First, we sample $n$ nonmember samples from LAION-mi. For each single sample, we finetune the Stable Diffusion v1.4 model on a single training step at $t=100$ using \textit{LoRA}. We measure the ratio of the training loss after and before the finetuning. We repeat that procedure for  $n$ member samples from LAION-mi. 
Thus we obtain loss ratio distributions for shadow models finetuned on member ($D_{mem}$) and nonmember ($D_{nonmem}$) samples. For inference, we fine-tune the model on the new sample using the same procedure, obtaining the corresponding loss ratio $l\star$. If $\frac{Pr[l\star|D_{mem}]}{Pr[l\star|{Dnonmmem}]} > \tau$ we classify the sample as a member. 

This attack can only be performed in the white-box scenario. The attacker needs to have access to the model weights to perform the fine-tuning.

We evaluate our method on 100 subsets of 1000 member and 1000 nonmember samples randomly selected from 4000 samples from each set. The approach based on shadow models achieves TPR@FPR=1\% equal to 2.21\% $\pm$ 1.11. The performance of the attack is thus comparable with the attacks described in Section~\ref{sec:attack_setting}. Although shifting the focus of shadow models from full training to finetuning using \textit{LORA} enabled us to apply this kind of approach, the resources required to execute this attack are still significantly larger than for the methods described in Sec.~\ref{sec:attack_setting}.

\section{Appendix: LAION-mi samples}
\label{sec:appendix_f}

In this section we provide a random sample of the LAION-mi dataset, 16 images from the nonmembers set and 16 images from the members set, see Figure \ref{fig:laion_mi_imgs}.

\section{Appendix: GPU cost}
To conduct the experiments for this paper we utilized $4000$ hours of Nvidia A100 GPU compute.

\end{document}